# CONTRACTUAL DEEPFAKES: CAN LARGE LANGUAGE MODELS GENERATE CONTRACTS?

Eliza MIK*

*Notwithstanding their unprecedented ability to generate text, LLMs do not understand the meaning of words, have no sense of context and cannot reason. Their output constitutes an approximation of statistically dominant word patterns. And yet, the drafting of contracts is often presented as a typical legal task that could be facilitated by this technology. This paper seeks to put an end to such unreasonable ideas. Predicting words differs from using language in the circumstances of specific transactions and reconstituting common contractual phrases differs from reasoning about the law. LLMs seem to be able to generate generic and superficially plausible contractual documents. In the cold light of day, such documents may turn out to be useless assemblages of inconsistent provisions or contracts that are enforceable but unsuitable for a given transaction.*

"What do you read, my lord?"
"Words, words, words."[1]

## Introduction

Can Large Language Models ("LLMs") generate contracts? Given their uncanny ability to write human-like stories, solve math problems and pass bar exams, the answer must be a resounding "yes"! Surely, something trained on the text of the entire Internet and capable of passing the Turing test,[2] must also be able to generate contractual provisions written in flawless legalese! It is beyond doubt that LLMs can *simulate* legal language and generate *seemingly* plausible contractual documents. Having "seen" millions of contracts, LLMs "know" what a contract should look like and can replicate popular contractual language well enough to generate "contractual deepfakes," documents that resemble contracts and even mimic the pleonastic style often preferred by lawyers. Whether such "contractual deepfakes" are or can serve as contracts is, however, a different question. Upon closer inspection, such documents may turn out to be contracts in form only, not in function. The preoccupation with text generation distracts from the fact that, in principle, the enforceability and commercial viability of a contract is not a question of form but of substance. Text generation is, however, not driven by reasoning or problem solving but by probabilistic word prediction. Statistical in nature, LLMs excel at computing relationships between words but remain oblivious to the relationships between words and the world. This paper opposes claims that LLMs will not only assist in contract review but also interpret[3] and

*Assistant Professor, Faculty of Law, The Chinese University of Hong Kong (elizamik@cuhk.edu.hk); Research on this paper has been supported by the CUHK Direct Research Grant No.4059053, titled "Testing the Limits of Machine Learning Approaches in the Automation of Legal Tasks: From Legal Prediction and Reasoning to Document Management and Generation."

[1] William Shakespeare, Hamlet, Act II, Scene ii.

[2] Cameron R. Jones; Benjamin K. Bergen, Large Language Models Pass the Turing Test (31 Mar 2025) arXiv:2503.23674v1

[3] see: Yonathan Arbel and David A. Hoffman, Generative Interpretation (2024) 99 NYU. L. Rev. X, 27-30, who propose that LLMs should answer legal questions concerning contractual interpretation; for a critique of this proposition, see: J Grimmelmann, B. Sobel, and D. Stein, Generative Misinterpretation (2025) Harvard Journal on Legislation 63 (1); Thomas R Lee and Jesse Egbert, 'Artificial Meaning?' (2025) 77 Florida Law Review; B Waldon et al., Large language models for legal interpretation? don't take their word for it (2026) Georgetown Law Journal 114.

generate entire contracts.[4] Instead, it aims to place the discussion on a firm factual footing.[5] The hype surrounding the purported capabilities of language models seems to derive from a general reluctance to engage with their technical underpinnings

Despite my obvious skepticism, the answer to the question whether LLMs can generate contracts requires a nuanced approach and some distinctions. . Pertinently, the answer will depend on our understanding of the word "contract," the minimum quality expected from a first draft and on our available resources. If contracts are regarded as *documents* filled with a *generic* collage of popular provisions, the answer is "yes." LLMs can generate probabilistic simulacra of contractual documents that could be adopted by uninformed parties or by those who simply do not care about the contents of the document. After all, many parties do not contemplate the possibility of litigation and sign arbitrary documents downloaded from the Internet as "representing their agreement." Often without reading. If, however, contracts are regarded as enforceable *agreements* addressing the commercial and legal aspects of *specific* transactions, the answer is "no." After all, the fact that the generated contract might be sufficiently certain and complete to be enforceable does not mean that it meets the needs of the contracting parties. Drafting contracts is not reducible to statistically plausible word predictions.[6]

## Outline of the Argument

Determining whether LLMs can generate contracts requires a baseline. To this end, the paper commences with a brief discussion as to what can be regarded as a "viable contract." The difficulty of establishing the minimum quality expected from a first draft – and the resources required to improve it – are a constant theme (or: *snag*?) in the analysis. Next, the paper introduces several distinctions regarding the term "contract." It must be remembered that contracts are *agreements* governing specific *transactions* and that generating text differs from recording the parties' agreement, facilitating transactions or allocating risks. Consequently, the ability to generate a *potentially* enforceable generic contract need not indicate the broader ability to generate a contract that can address the idiosyncrasies of concrete commercial exchanges. The paper proceeds with a description of LLMs. As my assertions that LLMs are only a sophisticated form of autocomplete[7] will not be taken on faith, I highlight three technical details that are generally underplayed: the difference between training and inference, the role of word embeddings and the probabilistic nature of text generation. I will demonstrate that (a) users of LLMs have relatively limited control over the text generation process, (b) LLMs only "know" how words relate to other words but lack any sense of the transactional context and (c) given the probabilistic nature of text generation, LLMs tend to replicate popular word patterns without regard to their potential effect on the rights and obligations of the contracting parties.

Next, the paper explains that the ability to generate text is unrelated to the ability to understand and use language. A brief interlude about a lonely (and bored!) octopus demonstrates the practical implications of the symbol grounding problem: knowing what words mean requires their association with objects and concepts in the real world. Understanding words requires physical embodiment and systems that have only ever experienced text, cannot understand context and apply the law to address concrete problems. The paper also examines whether LLMs can "know" the law. To address the misconception that knowledge can emerge from petabytes of data, it introduces the concept of parametric knowledge and distinguishes between *knowing* and *applying* the law as well as between memorizing word patterns and reasoning about the law. Anticipating objections that several techniques could, theoretically, enable

[4] Contract Intelligence and Rise of Large Language Models in Contract Management (June 5, 2024) Shibhu Nambiar, Senior Director at Icertis, https://www.icertis.com/learn/how-do-large-language-models-work-in-contract-management/ R. Gainer, K. Starostin, 'How LLMs Can Boost Legal Productivity (with Accuracy and Privacy)' Apr 2024. URL https://cohere.com/blog/how-llms- can-boost-legal-productivity-with- accuracy-and-privacy; see: H. James Wilson and Paul R. Daugherty, Embracing Gen AI at Work (September-October 2024) HBR 151 who state that "according to our research, most business functions and more than 40% of all U.S. work activity can be augmented, automated, or reinvented with gen AI. The changes are expected to have the largest impact on the legal, banking, insurance, and capital-market sectors;" but note that A recent study found that AI-based tools improved the performance of legal professionals in a number of litigation-oriented tasks; but did not appear to extend to contract drafting, see: Daniel Schwarcz et al., AI-Powered Lawyering: AI Reasoning Models, Retrieval Augmented Generation, and the Future of Legal Practice (March 02, 2025). Minnesota Legal Studies Research Paper No. 25-16, Available at SSRN: https://ssrn.com/abstract=5162111 at 8

[5] Abeba Birhane et al., "Science in the Age of Large Language Models" (2023) 5 Nature Reviews Physics 277-80.

[6] Thomas McCoy et al., *Embers of Autoregression: Understanding Large Language Models Through the Problem They are Trained to Solve*, ARXIV (Sep. 24, 2023), https://arxiv.org/pdf/2309.13638.pdf. 3, 4

[7] Arvind Naraynan, Sayash Kapoor, AI Snake Oil (Princeton University Press, 2024) 132

LLMs to generate viable contracts, the paper discusses the possibility of training LLMs on legal texts as well as improving their performance with additional information provided at inference. In the final part, the discussion examines the risks of signing (or otherwise adopting) generated documents that look like contracts but, in most instances, are unenforceable, do not reflect the parties' agreement or lead to commercially nonsensical outcomes. It also compares the resources required to review or "repair" generated contracts to those involved in the customization of templates.

I reserve *drafting* to traditional methods of preparing contractual documents and assume that lawyers do not produce words but apply the law and solve problems. In most instances, contractual documents comprise numerous, implicit confrontations with legal questions that arise in the context of particular transactions. I also assume that contract drafting requires knowledge of the world and the ability to reason. I bracket the question whether LLMs can transform legal practice in general.

## Setting the Baseline

There is no "perfect contract." The quality of a contract is a matter of opinion and competence. For present purposes, the term "viable contract" is used to indicate a contractual document that is *adequate* in light of the parties' transaction and *compliant* with the jurisdiction-specific legal requirements. Regarding adequacy: contracts of sale should not, for example, contain provisions specific to leases and construction contracts should not contain provisions tailored to charterparties. Regarding compliance: apart from meeting the requirements of enforceability, contracts must be drafted with the relevant laws in mid. In England and Wales, for example, contracts cannot be formed for illegal purposes or contravene the rule against penalties. Viablity can also depend on the context: a contract between sophisticated businessmen may be unsuitable in a consumer setting. In principle, a viable contract must "fit" the transaction *and* the applicable law. Enforceability could thus be regarded as the bare minimum, whereas the reconciliation of the parties' conflicting interests in light of their commercial goals, bargaining power and prevailing legal framework, the ideal. The first (often superficial) indication of viability is the parties' willingness to adopt the contractual document as expressing their agreement. Yet, if neither party read the document and/or has a limited understanding of the law, they may sign *any* document written in legalese. Superficial plausibility is often mistaken for substantive adequacy, especially by non-lawyers. A better indication of viability would be if the parties' lawyers concluded that the generated document fulfilled their clients' needs. Admittedly, the viability of any contract can only be established when the parties commence performance or, in the event of a dispute, after its contents are evaluated by an adjudicator. Unlike code, which is increasingly generated by LLMs, contracts cannot be "tested and run" before they are signed. Viability can also be associated with a document that is "worth editing." If, however, the generated contract is regarded as a draft, then the question is not whether LLMs can generate contracts but whether it is efficient to use them for that purpose. If more resources are required to review and to substantively rewrite the generated document than to draft it in a traditional manner, the answer to the question whether LLMs can generate contracts cannot be affirmative.[8]

It is counter-productive to inquire whether LLMs can generate better contracts than humans. Everything depends on the human in question and what is considered "better." Even contracts drafted by lawyers may be imperfect due to conflicting aims, simple errors, differing drafting practices or hasty compromises.[9] It is also unclear how to measure performance improvements when it comes to drafting. Instead of comparing traditional contracts to their synthetic cousins, we should ask: what are the main challenges when drafting? Typing speed aside, is it the difficulty of minimizing the potential of disputes? The difficulty of drafting with precision? Or is it something more trivial: reducing the time needed to review initial drafts? Throughout the discussion, two questions should be kept in mind: "what problem can be solved by LLMs in the context of drafting contracts?" and "is it more efficient to generate a new document or to customize an existing document?" LLMs only reduce the time needed to create text, not the time required to read text.

[8] Yuvaraj, Joshua, The Verification-Value Paradox: A Normative Critique of Gen AI in Legal Practice (October 18, 2025). The University of Auckland Faculty of Law Research Paper Series 2026, (2026) 52 *Monash University Law Review* (forthcoming), Available at SSRN: https://ssrn.com/abstract=5621550 or http://dx.doi.org/10.2139/ssrn.5621550

[9] *Wood v Capita Insurance Services Limited* [2017] UKSC 24 at 13

## A Contractual Primer

The answer to the question whether LLMs can generate contracts depends on our understanding of the term "contract." We must distinguish between contracts in the sense of *abstract agreements*, and contracts in the sense of *physical documents*, as well as between the *formation* of contracts and the *preparation* of contractual documents. We must also recall that the function of *a* contract is, amongst others, to "determine the precise legal effect of that agreement"[10] and to provide legal certainty that each party will obtain what he or she bargained for[11] or, at least, be able to ask the court for assistance if the other fails to perform. Contracts are not pages covered in legalese, but *agreements* made to support *specific* transactions.[12] The techno-enthusiasm surrounding LLMs obscures the fact that people form contracts to achieve commercial results, not to produce documents. As contracts need not be made in writing[13] and as written evidence of a contract is required in limited instances,[14] the preparation of contractual documents is optional. Contracts are put in writing mainly for evidentiary reasons, to facilitate enforceablity, [15] particularly in high-value or high-risk transactions. Contractual documents serve a purpose; form is subservient to function. The following distinctions are thus crucial:

*First*, the term "contract" can refer to the parties' *agreement* and to the *document* describing such agreement. The contractual document is not, however, synonymous with the agreement.[16] The medium used to record the contract constitutes proof of the parties' obligations, but it does not *create* them. The source of rights and obligations is the agreement (in the abstract sense). There can be no contract without agreement, but there can be a contractual document that does not record anyone's agreement. Examples are contractual templates prepared for use in future dealings, standard form contracts awaiting adoption as well as contractual documents reflecting complex negotiations that have been abandoned if one or both parties withdrew from the transaction. Contractual documents that have not been signed or otherwise adopted by the parties are only pieces of paper covered in words.[17]

*Second*, the *creation* of a contractual document differs from the *formation* of a contract. There is no prescribed sequence as to what comes first but the fact that the parties have reached agreement, does not mean that they have or will put it in writing. Conversely, the fact that a contractual document has been drafted does not mean that a contract has been formed. Contractual documents can be prepared in advance or, as indicated, abandoned if the parties decide not to proceed.

Lastly, parties often sign documents they have not read, do not understand and that cannot be said to represent their agreement. And yet, following the objective theory of contract, it will be assumed that the document *is* the agreement.[18] The objective theory of contract is indifferent to the technical provenance of the text of the contract. Irrespective of whether such text is typed, handwritten or computer-generated, it represents an outward manifestation of the parties' common intention. For reasons of commercial convenience, the "parties are judged, not by what is in their minds, but by what they have said, written or done."[19] Discussions concerning the feasibility of "contract generation" will always be overshadowed by the cynical observation that contracts are rarely read and that few contractual documents reflect the parties' *actual* agreement. Moreover, if a contract is performed without problems - and without reference to the document - then its contents may in fact be regarded as irrelevant. If each party performs to the satisfaction of the other, the contractual document may never be consulted. Its contents become important, however, in the event of disputes. In other words, the written agreement becomes important in case of disagreement.[20] Suddenly, every word matters, and the document is meticulously dissected to "discover" what exactly had been agreed. As demonstrated below, the fact that the parties decided to sign a document that *looks like a contract* does not mean that it *is* a contract or that it fulfils *their* commercial and legal needs. Moreover, the fact that the generated document could be an enforceable contract does not mean that it is a viable contract.

[10] Edwin Peel, *Treitel On the Law of Contract* (15th edn Sweet and Maxwell 2020) para 1-009
[11] J Beatson, J Burrow and J Cartwirght, *Anson's Law of Contract* (31st edn OUP 2020), 3, 4
[12] Edwin Peel, *Treitel On the Law of Contract* (15th edn Sweet and Maxwell 2020) para 1-001; J Beatson, J Burrow and J Cartwright, *Anson's Law of Contract* (31st edn OUP 2020), 2.
[13] H. Beale (gen. ed.) *Chitty on Contracts*, 35nd edn (Oxford: Sweet & Maxwell, 2024) at para. 1-065; J. Beatson, A. Burrows, J. Cartwright, *Anson's Law of Contract*, 31th ed. (Oxford: Oxford University Press, 2010) at p. 77.
[14] see, for example, the Law of Property Act 1925, ss 52, 54 (2); Law of Property (Miscellaneous Provisions) Act 1989, ss 1(2), (3).
[15] Hugh G Beale and others (eds), *Chitty on Contracts* (35th edn Sweet and Maxwell 2024), para 8-001
[16] D. Koepsell and B. Smith, 'Beyond Paper' (2014) 97(2) The Monist 222; P. Amselek, 'Philosophy of Law and the Theory of Speech Acts' (1998) 1(3) Ratio Juris 187.
[17] Yet, the fact that a contractual document was adopted does not automatically mean it is a contract!
[18] Patric Atiyah, *Essays on Contract* (OUP 1990) 21; *FSHC Group Holdings Ltd v Glas Trust Corporation Ltd* [2019] EWCA Civ 1361 at 149
[19] See Chitty 1-053 citing *Furmston, Cheshire, Fifoot and Furmston's Law of Contract, 17th edn (2017), p.42.*

Once our attention shifts from documents and text to agreements and transactions, the idea of resorting to LLMs starts loosing its appeal. Especially if we consider their technical characteristics.

## The Technological Background

LLMs are statistical models of word distributions in a language.[21] This means that they can replicate, or mimic, common word patterns specific to a language. Crucially, text generation relies on word prediction. When given a prompt, the operation of LLMs such as OpenAI's GPT series or Google's Gemini family can be described as follows: "Here's a fragment of text. Tell me how this fragment might go on. According to your model of the statistics of human language, what words are likely to come next?"[22] Generating their response word by word,[23] LLMs first compute the most probable (or, depending on the settings, a reasonably probable)[24] next word wn+1, then the word (wn+2) that is most likely to follow (w1 . . . wn+1). LLMs repeatedly compute the most probable next word, until they reach a preset word limit.[25] When calculating probabilities, LLMs do not make decisions based on the substance of the text in the prompt or the generated text. In fact, they do not engage with the meaning of words at all as the word distributions learned by LLMs are "statistically independent" of their semantic content.[26] Text generation is driven *exclusively* by the conditional probability of producing word wn following the probability function P (wn |w1...wn−1). To fully convey the limitations of LLMs, three technical aspects of word generation require emphasis.

### Training vs Inference

As LLMs are first *developed* and then *deployed*, we must distinguish between *training* and *inference*. This distinction is often underplayed, leading to misunderstandings concerning the degree to which LLMs can be steered and, more importantly, their inherent isolation from the circumstances in which they are deployed. In principle, users have no control over the development (training) of LLMs[27] and relatively limited control over their deployment (inference). Training LLMs generally involves two steps. During *pre-training*, models are provided with enormous amounts of raw text, the so-called training corpus, from which they learn the statistical characteristics of language.[28] Next, most LLMs undergo *post-training*, which generally involves finetuning on labelled examples and/or human evaluations of the generated output.[29] While pre-training relies on brute-force statistics, post-training is more task-oriented and aims to align model performance with concrete user preferences. Once training is completed, LLMs do not evolve further. Their "knowledge" of word distributions is static and does not dynamically adapt to the circumstances in which they are deployed. At inference, LLMs generate text in response to written user instructions, known as prompts. Prompts can *to an extent* steer text generation towards specific tasks. Their impact on the quality of the generated output is, however, relatively limited as models often ignore or "misinterpret" the prompt and fall back on the word patterns learned during training.

---

[20] R. Scott, 'A Relational Theory of Default Rules for Commercial Contracts' (1990) 19 Journal of Legal Studies 597.
[21] McCoy et al (n 6)
[22] Murray Shanahan, Talking about large language models (2024) Communications of the ACM 67 (2) 70
[23] To be more precise: the basic prediction units are *tokens*, elements of text, such as words and fragments of a word.
[24] LLMs have different "temperature" settings. The lower the temperature, the more deterministic its output. Higher temperatures introduce more randomness and make outputs more "creative" but also more unpredictable.
[25] Davis (n ) 289, 290.
[26] Adam Bouyamourn, Why LLMs Hallucinate, and How to Get (Evidential) Closure: Perceptual, Intensional and Extensional Learning for Faithful Natural Language Generation in: Proceedings of the 2023 Conference on Empirical Methods in Natural Language Processing, pages 3181–3193
[27] Training LLMs requires significant amounts of GPUs, the cost which is prohibitive to most companies, not to mention individual users.
[28] Alec Radford et al., *Improving Language Understanding by Generative Pre-Training* (2018) https://cdn.openai.com/researchcovers/languageunsupervised/language_understanding_paper.pdf.
[29] Long Ouyang et al., *Training Language Models to Follow Instructions with Human Feedback*, 36 PROC. CONF. ON NEURAL INFO. PROCESSING SYS. (2022).

## Word Embeddings

Although LLMs generate words, they can only process numbers. Their ability to perform computations on text is attributable to so-called word embeddings i.e., numerical representations of words that capture many of their linguistic properties based on their relationship with other words.[30] Word embeddings are represented in multi-dimensional vector spaces, with each word occupying one point in such space and each dimension capturing some aspect of its meaning.[31] Words with similar meaning or words frequently occurring together are said to be close in the vector space.[32] For example, the embeddings for 'contract' and 'agreement' are closer than those for 'contract' and 'ship.' As embeddings reflect word proximity (e.g. words in popular phrases) and not necessarily semantic similarity, even linguistically unrelated words can have similar word vectors.[33] 'Sign' may thus be close to both 'contract' and to 'street' given the common word sequences 'to sign a contract' and 'street sign.' Word embeddings enable LLMs to calculate whether the word "bank" relates to a river or to a financial institution. What is important for the present discussion is that word embeddings represent statistical relationships between words in *specific* training corpora but not, as a contract lawyer might assume, a word's "natural and ordinary" meaning.[34] As word embeddings depend on the text used in training,[35] LLMs trained on social media posts will feature different relationships between words than LLMs trained on cases on contract law. Although word embeddings reflect how words are used in the myriad of contexts captured in their training corpora,[36] they are unrelated to the context in which LLMs generate text at inference. The implications are easy to miss: word embeddings provide invaluable insights into *general* word usage, as captured in their training data, but contract law is only concerned with how a word (or word sequence) is used in the *single* context in which a document is created.[37] Moreover, the fact that the meaning of words in a contract is evaluated objectively, does not imply that such meaning is static, generic or universal.[38] Quite the opposite. Meaning is transaction-specific as every contract has its own "back story," including the facts and circumstances known or assumed by the parties *when it was made*.[39] To re-emphasize: words in a contract derive their meaning from the document and from the factual matrix of the transaction in which it is created.[40] Not from multidimensional vector spaces that reflect relationships between words in terabytes of arbitrary training text. In contract law, what matters is the time when the LLM is deployed (inference). The "reasonable person" is, after all, deemed to have all the background knowledge, which "includes absolutely anything which would have affected the way in which the language of the document would have been understood" by the parties *at the time of contracting*.[41] Word embeddings, however, relate to the time when the LLM was developed (training).

Word embeddings are the substrate for the so-called attention mechanism, which enables LLMs to select (i.e. pay *attention* to) the most salient parts of the input and to capture long-range dependencies

[30] Omer Levy and Yoav Goldberg, Linguistic Regularities in Sparse and Explicit Word Representations (2014) in: Proceedings of the Eighteenth Conference on Computational Natural Language Learning, pages 171–180, Ann Arbor, Michigan. ACL

[31] Dimensions denote the number of features representing data points within a dataset. For example, a two-dimensional space requires two parameters (e.g., length and width) to define the position of a point, while a three-dimensional space adds another parameter, such as height. In practice, each word is associated with a vector in the embedding space.

[32] Each dimension represents an attribute, and each word is represented as a unique combination of these attributes, a simplified word vector for 'cat' might have a high value for the attribute 'animal', an intermediate value for 'pet' and 'small', and a low value for 'vehicle;' for a simple explanation see: Yonathan Arbel and David A. Hoffman, Generative Interpretation (2024) 99 NYU. L. Rev. 1, 26, 27; the use of statistical modelling to capture meaning dates back to 1957, when British linguist Firth asserted "you shall know a word by the company it keeps" sought to approximate meaning through the representation of shared contexts; see: John R Firth, "A Synopsis of Linguistic Theory, 1930-1955" (1957) Studies in Linguistic Analysis

[33] see: Omri Ben-Shahar & Lior J. Strahilevitz 'Interpreting Contracts via Surveys and Experiments,' (2017) 92 N.Y.U. L. REV. 1753; Farshad Ghodoosi & Tal Kastner, 'Big Data on Contract Interpretation' (2024) U.C. DAVIS L. REV. 1, 58

[34] for an attempt to use LLMs to determine the ordinary, everyday meaning of words, see: *Snell v. United Specialty Ins. Co.*, Civil Action 21-0229-CG-M (S.D. Ala. Jul. 6, 2022) Newsom J, 4, 12, 14

[35] Jonathan H. Choi, Measuring Clarity in Legal Text (2024) 91 U. Chicago L. Rev. 1, 46-48.

[36] Choi, (n X) 20

[37] *Rainy Sky SA v Kookmin Bank* [2011] UKSC 50 at 14-30; *Arnold v Britton* [2015] UKSC 36 at 14-22; and *Wood v Capita Insurance Services Limited* [2017] UKSC 24 at 8-15; Lord Hoffmann in *Chartbrook Ltd v Persimmon Homes Ltd* [2009] UKHL 38, [2009] 1 AC 1101, para 14; *Investors Compensation Scheme Ltd v West Bromwich Building Society* [1998] 1 WLR 896 Lord Hoffmann (pp 912-913).

[38] S. Macaulay, 'The Real and the Paper Deal: Empirical Pictures of Relationships, Complexity and the Urge for Transparent Simple Rules' (2003) 66 MLR 44, 45

[39] Arnold Lord Neuberger at [15], 19, 20; for this very reason precedents on interpretation are of limited value, because "a decision on a different clause in a different context is seldom of much help on a question of construction, *Surrey Heath BC v Lovell Construction Ltd* (1990) 48 Build LR

[40] *Wood v Capita Insurance Services Ltd* [2017] UKSC 24, [2017] AC 1173, para 10; *Pink Floyd Music Ltd and another v EMI Records Ltd.* [2010] EWCA Civ 1429 at 17, 18; *Savills (UK) Ltd v Blacker* [2017] EWCA Civ 68, [33]

[41] *Chartbrook Ltd v Persimmon Homes Ltd* [2009] UKHL 38, [2009] 1 AC 1101, para 14

between words in preceding paragraphs or even pages.[42] While attention enables LLMs to generate long passages of coherent text, it does not change their autoregressive nature: text is generated one word at a time, with each newly generated word conditioned on previous words.[43] This "left-to-right" generation means that (a) the text cannot be retroactively adjusted or corrected, and (b) LLMs cannot plan ahead "what they are going to say overall."[44] This may lead to an accumulation of errors and inconsistencies, especially when models are prompted to generate longer texts. Lacking a "coherent picture of the overall response,"[45] LLMs cannot "think of" a document as a whole and anticipate the dependencies between the text they already generated and the text that will – *or should be* - generated next.[46] In complex, multi-step generation tasks LLMs tend to gradually lose coherence or deviate from the original instructions.[47] A generated provision may seem plausible when read in isolation. When read in context of the entire document, however, it may conflict with other provisions, creating unexpected ambiguities and inconsistencies.[48] To recall: when interpreting a contract, courts first examine the document *as a whole*,[49] reading individual provisions in light of *all* other provisions.[50] Contractual documents often contain cross-references, qualifications and other long-range syntactic dependencies between different terms spanning the entire document.[51] Recitals and exclusion clauses co-exist with indemnities and liquidated damages, weaving hierarchical webs of interrelated provisions that affect the interpretation of individual words. Contract drafting is an iterative process of ensuring both structural and semantic coherence.

## Probabilistic Nature

LLMs are inherently probabilistic and predict the next word based on how likely such word is to appear after a previous word(s). Provided with the prompt "my favorite color is __", an LLM should predict "blue" (or some other color) rather than "coconut" or "contract." An indication of colour is, after all, more likely than any of the aforementioned nouns. As word probabilities are calculated based on the word distributions learned during training, word sequences that frequently occur in the training corpora are naturally more probable. Consequently, LLMs tend to favour high-probability words and reproduce word strings or even entire phrases they have frequently encountered during training.[52] Although there is no one-to-one mapping from the training text to the generated text, some outputs may be exact copies of common phrases; others are best described as "recombinations" of popular word strings. Whether the word prediction process should be described as "reproduction," "regurgitation" or "recombination" is subject to debate. What matters in practice is that, when generating text, LLMs tend to repeat popular word patterns.[53] Unsurprisingly, LLMs fare better at tasks which require the generation of frequent word combinations and cannot reliably deal with tasks requiring word combinations that significantly differ from those common in their training corpora.[54] While the exact contents of such corpora remain undisclosed,[55] often due to fear of copyright repercussions,[56] they are known to comprise large swaths

[42] Ashish Vaswani et al. *Attention is All you Need*, 31 CONF. ON NEURAL INFO. PROCESSING SYS. (NIPS 2017)
[43] See generally: Emily M. Bender et al., On the Dangers of Stochastic Parrots: Can Language Models Be Too Big? ACM CONF. ON FAIRNESS, ACCOUNTABILITY, & TRANSPARENCY 610, 616–17 (2021).
[44] Ernest Davis, 'Mathematics, Word Problems, Common Sense, and Artificial Intelligence,' (2024) 61 Bulletin (New Series) of the American Mathematical Society 287, 292; Chris Su and others, 'Limits of Emergent Reasoning of Large Language Models in Agentic Frameworks for Deterministic Games' (arXiv, 12 October 2025) doi:10.48550/arXiv.2510.15974.
[45] Naraynan and Kapoor (n 7) 133
[46] Yulong Wu et al., 'Natural Context Drift Undermines the Natural Language Understanding of Large Language Models' (arXiv preprint, arXiv:2509.01093, 1 September 2025) https://arxiv.org/pdf/2509.01093.pdf accessed 30 October 2025; Vardhan Dongre et al, 'Drift No More? Context Equilibria in Multi-Turn LLM Interations' (arXiv preprint, 9 October 2025) arXiv:2510.07777v1 https://arxiv.org/pdf/2510.07777v1.pdf
[47] Yulong Wu et al.,
[48] Peter V Coveney and Sauro Succi, 'The wall confronting large language models' (2025) arXiv:2507.19703 https://arxiv.org/abs/2507.19703 accessed 3 November 2025.
[49] *Arnold v Britton* [2015] UKSC 36, Lord Neuberger at [15]; *Wood v Capita Insurance Services Ltd* [2017] UKSC 24 para X; Zhong Xing Tan, 'Beyond the Real and the Paper Deal: The Quest for Contextual Coherence in Contractual Interpretation' (2016) 79 MLR 623, 643.
[50] *Pink Floyd Music Ltd and another v EMI Records Ltd.* [2010] EWCA Civ 1429 at [x].
[51] Eric Martinez et al., 'Even Lawyers do not Like Legalese' (2023) 120 PNAS 1
[52] McCoy 20, 23, 24; B.Y. Lin et al., 'Birds have four legs?! Numbersense: Probing numerical commonsense knowledge of pre-trained language models' In: Proceedings of the 2020 Conference on Empirical Methods in Natural Language Processing (EMNLP). pp. 6862–6868 (2020)
[53] Naraynan and Kapoor (n 7) 136
[54] Zhaofeng Wu, Reasoning or Reciting? Exploring the Capabilities and Limitations of Language Models Through Counterfactual Tasks (28 March, 2024) https://arxiv.org/abs/2307.02477
[55] Ernest Davis , 'Mathematics, Word Problems, Common Sense, and Artificial Intelligence,' (2024) 61 Bulletin (New Series) of the American Mathematical Society 287
[56] Shayne Longpre et al., Consent in Crisis: The Rapid Decline of the AI Data Commons (2024)

of online content, including terabytes of Reddit posts, marketing materials and news articles.[57] The problem is not, however, that the word sequences common in such content cannot be regarded as adequate building blocks for the generation of contracts, or that contracts and legal sources are statistically underrepresented.[58] The problem is that the adequacy of a contractual provision depends on the circumstances of the transaction – not on the probability of a word! Temporarily bracketing questions of knowledge and reasoning, a viable contract may require sequences of words that are uncommon or altogether absent from the training corpora. As probabilistic models favour frequency, LLMs are, in principle, less reliable in domains with idiosyncratic word distributions,[59] such as those encountered in contracts. Pertinently, even if the training corpora contained a statistically significant number of contractual documents covering a wide variety of transactions, LLMs would still operate at the level of word distributions, not at the level of substantive provisions. They would remain unable to *analyze* which words should be included in contracts governing specific transactions or to generate provisions addressing *novel* legal developments. There is simply no "sufficient" number of contracts that an LLM could be trained on for it to be able to generate viable contracts. To aggravate matters, LLMs seem to favor content generated by other LLMs over that created by humans[60] and, if trained on low-quality text (e.g. sensationalistic media content), provide increasingly incomplete or superficial answers.[61] Such technicalities might, again, seem of limited relevance were it not for the fact that LLMs are trained to generate text irrespective of their familiarity with a particular task.. This often leads to so-called hallucinations: superficially plausible and seemingly relevant but incorrect or nonsensical outputs.[62] Statistically probable does not, after all, mean correct or adequate.[63]

In sum, when generating text, LLMs rely on statistics not semantics. Word embeddings reflect word usage learned during training but are detached from the circumstances in which a contractual document is generated. Driven by probability, LLMs tend to replicate word patterns frequently occurring in their training corpora. Their performance deteriorates when prompted to generate text that they had rarely or never encountered. Despite these shortcomings, LLMs never refuse to respond to a prompt[64] and confidently recombine word sequences occurring in *popular* contractual provisions regardless of their suitability for a particular transaction. To the untrained eye, however, the generated text will look adequate. Its superficial plausibility will discourage diligent review.

### Understanding and Symbol Grounding

LLMs do not understand words.[65] In a seminal 2022 article, Bender and Koller demonstrate that the ability to understand language cannot, as a matter of principle, be learned solely from text.[66] For LLMs to understand words, it would be necessary to solve the perennial symbol grounding problem,[67] that is, to train them to associate arbitrary symbols (words) with concrete referents (objects, concepts and actions).[68] There is, after all, no natural association between words and objects: "coconut" is not a universal signifier of the fruit. The symbol grounding problem gains in complexity with abstract referents, like "indemnity" or "delivery," which require associations with concepts.[69] Even then, however, such associations require *prior* internal representations that exist independently from text and, at some stage, require grounding in the physical world. Training LLMs on thousands of cases containing the *word* "delivery" would thus be pointless, because understanding the concept of "delivery" requires familiarity with "possession" and "control." It also requires basic knowledge of physical objects, (e.g. coconuts cannot exist in two places simultaneously) as well as spatial relationships, (e.g. the delivery of coconuts by sea takes significantly longer than by air.) At present, LLMs remain unable to appreciate such relationships, not to mention grasp the concepts of time and distance. They only "know" which words often occur together with – or in proximity of – "deliver." To clarify: Bender and Koller are not the first to observe that understanding requires that words be associated with "*something*" and that such associations are learned through embodied experiences.[70] *Nihil in intellectu nisi prius in sensu*: there is nothing in the intellect that was not first in the senses! The idea that language derives from embodied activity has been expressed in several disciplines, including philosophy,[71] linguistics[72] and robotics![73] In sum: contracts are made in the physical world and the meaning of words derives from the circumstances of a transaction.

---

[57] The latest crawl contained 2.7 billion web pages, while the previous crawl contained 3.1 billion, see: https://www.commoncrawl.org/blog/june-2024-crawl-archive-now-available; see also: Stefan Baack, A Critical Analysis of the Largest Source for Generative AI Training Data: Common Crawl. In Proceedings of the 2024 ACM Conference on Fairness, Accountability, and Transparency (FAccT '24). ACM, New York, NY, USA, 2199–2208.

[58] It is estimated that only about 4% of such corpora pertain to law and government, see: Chip Gyuen, AI Engineering, O'Reilly 2024 at X

[59] Richard Heersmink et al., A phenomenology and epistemology of large language models: transparency, trust, and trustworthiness (2024) 26 Ethics and Information Technology 41

[60] W. Laurito et al., AI–AI bias: Large language models favor communications generated by large language models, Proc. Natl. Acad. Sci. U.S.A. 122 (31) e2415697122, https://doi.org/10.1073/pnas.2415697122 (2025).

[61] Shuo Xing and others, 'LLMs Can Get "Brain Rot"!' (arXiv preprint, 14 October 2025) https://arxiv.org/abs/2510.13928 accessed 27 October 2025

[62] Pranav Narayanan Venkit, et al., "Confidently Nonsensical?": A Critical Survey on the Perspectives and Challenges of 'Hallucinations' in NLP (11 April, 2024)

[63] Gary Marcus, *Taming Silicon Valley* (MIT Press 2024) 165

[64] Except in situations where a particular prompt triggers the moderation mechanisms of the LLMs creator, as would be the case if a user requested advice on committing a crime etc.

[65] While the general consensus in technical literature is that LLMs do not understand text, some controversies remain, see: Mitchell M., Krakauer D.C. 'The debate over understanding in AI's large language models' (2023) Proceedings of the National Academy of Science 120(13); see: Melanie Mitchell 'AI's challenge of understanding the world' (2023) 382 Science 8175; appearances of understanding based on successful benchmark performance reveal that LLMs cannot use the concepts they purportedly understand, see: Marina Mancoridis et al., Potemkin Understanding in Large Language Models, Proceedings of the 42 nd International Conference on Machine Learning, Vancouver, Canada. PMLR 267, 2025.

[66] Emily M Bender and Alexander Koller, 'Climbing towards NLU: On Meaning, Form, and Understanding in the Age of Data' (2020) Proceedings of the 58th Annual Meeting of the Association for Computational Linguistics 5185–5198

[67] Stevan Harnad, The symbol grounding problem (1990) 42 Physica D: Nonlinear Phenomena, 335-346, Selmer Bringsjord, 'The symbol grounding problem… remains unresolved' (2015) *Journal of Experimental & Theoretical Artificial Intelligence* 27:1, 63-72.

## An Interlude: The Lonely Octopus

Bender and Koller illustrate the impossibility of learning meaning from text with a tale of a lonely octopus and two shipwrecked humans. A and B, two English speakers, are independently stranded on two uninhabited islands.[74] Mysteriously, previous inhabitants have left telegraphs so that A and B can exchange messages via an underwater cable. Battling their boredom, A and B engage in long conversations. O, a hyper-intelligent octopus living in the azure waters surrounding the islands, taps into the cable and starts eavesdropping on their communications. While O does not understand any of the words exchanged by the parties, it does excel at detecting patterns. Gradually, by observing the parties' communications or, more specifically, the word patterns in their messages, O learns to predict how B will respond to A. "O also observes that certain words tend to occur in similar contexts, and perhaps learns to generalize across lexical patterns by hypothesizing that they can be used somewhat interchangeably."[75] Although O cannot understand, it can identify sequences of common word patterns. One day, feeling particularly lonely, O cuts the cable and inserts himself into the conversation, by pretending to be B. From that moment, all communications occur between O and A. B succumbs to solitary despair. Can O pose as B without making A suspicious? Can O maintain a human-like conversation? The answer depends on the task and on the purpose of a given communication. As long as A wants to converse about their daily lives, O may be able to "generate" word-strings of the kind produced by B because such "conversations have a primarily social function, and do not need to be grounded in the particulars of the interlocutors' actual physical situation nor anything else specific about the real world."[76] It suffices that O produces text that is internally coherent and follows similar patterns to those previously exchanged. In short, O can replicate prior conversations on the basis of his knowledge of word distributions. One day, A invents a coconut catapult and sends B (that is, O) instructions how to build the apparatus. O, however, cannot build the catapult because it does not understand *what* "catapult" and "coconut" refer to, or *mean*. O has never left the sea. O knows word distributions but it does not know the world. Imitating B's prior responses to similarly worded text, O responds: "Cool idea!" A may accept this response as satisfactory — but only because A attributes meaning to O's message and does not expect O to immediately build the catapult. One day, A falls ill and describes her symptoms to B (that is, O). In such situation, A asks for help and wants O to *solve a problem*. O, however, has never encountered a similar word pattern. As an adequate response would require understanding, not to mention reasoning, O remains silent. A perishes. Having only experienced text, O did not learn meaning. Irrespective of the size of the training corpus (the number of messages O has seen) and the length of training (the time O monitored the communications), LLMs trained exclusively on text cannot learn to associate words with objects and concepts. Just like O, LLMs cannot grow little legs and explore the world.

[68] Morten H. Christiansen, Nick Carter, *The Language Game* (Penguin Random House, 2022) 59
[69] Steven T. Pintadosi & Felix Hill, *Meaning without Reference in Large Language Models,* ARXIV (Aug. 12, 2022) https://arxiv.org/pdf/2208.02957.pdf.
[70] J. McCarthy and P.J. Hayes, "Some Philosophical Problems from the Standpoint of Artificial Intelligence," in: B. Meltzer and D. Michie, eds., *Machine Intelligence*, vol 4 (Edinburgh: Edinburgh University press, 1969) 463, 468, 469; Sagnik Ray Choudhury et al., Machine Reading, Fast and Slow: When Do Models "Understand" Language? (2022) 29th ICCL 78-93; Floridi (n X) 137.
[71] Ludwig Wittgenstein, *Philosophical Investigations* (Oxford: Blackwell, 1953) 4th edition, 2009, P.M.S. Hacker and Joachim Schulte (eds. and trans.), Oxford: Wiley-Blackwell, 43, 220.
[72] C.K. Ogden, I.K. Richards, The Meaning of Meaning (1923)
[73] Rodney A. Brooks, "Intelligence without representation, (1991) 47 Artificial intelligence 139
[74] Bender & Koller (n x)
[75] Bender & Koller (n x)
[76] Bender & Koller (n x)

Could O be tasked with drafting a contract for the sale of coconuts? Or a contract for the construction of a catapult? Can such contracts be drafted without *any* world experience? Can such experience be contained in (or: conveyed by?) the training corpora? Can the world be captured in text? To draft viable contracts for the sale of coconuts, lawyers must not only understand what coconuts are but also know their attributes. Will they decay without refrigeration? Disintegrate when dropped? Are they easily replaced or unique? LLMs "know" that "coconut" is close in the vector space to "milk" and "oil." They do not know, however, that (unlike catapults) coconuts are small and perishable. LLMs do not understand the concepts of "transport" or "delivery," not to mention the intricate relationships between payment, the transfer of ownership and the allocation of risk. Without an understanding of the contractual subject matter, LLMs do not know whether the contract would be goverened by the Sale of Goods Act 1979 or whether they were dealing with a *sui generis* transaction that requires idiosyncratic provisions.[77] To LLMs, catapults and coconuts are words in vector spaces, not physical objects.. LLMs can, unquestionably, generate a generic contract of sale by replicating word patterns commonly enountered in contracts of sale. LLMs would, however, struggle to generate a contract tailored to the sale of specific goods in specific circumstances - such as shipping coconuts from isolated islands! Such contract requires not only understanding but the rudimentary ability to anticipate problems and to plan ahead. Admittedly, the attributes of the subject matter, alongside with descriptions of transport arrangements, etc. could be provided in the prompt. As described below, the amount of text required to convey the indispensable details (including such obvious facts that shipping containers are better stored below deck!) may exceed the amount of text of the final contract. Crucially, lacking the ability to understand words, LLMs would remain be unable to grasp their practical implications. "Store below 30 C," would be a *word string* affecting the calculation of the next word, not *information* about storage conditions. While comprehensive prompts may, in theory, increase the probability of generating some viable contractual provisions, LLMs will always encounter the same obstacle: not understanding what words mean. The replication of word patterns, no matter how sophisticated, cannot replicate understanding.

## Generating vs Using Language

Despite their name, LLMs are *text* generators. Their inability to understand language translates into the inability to use language. The latter exceeds the production of semantically plausible utterances.[78] Technical scholarship distinguishes between *formal* linguistic competence, which denotes knowledge of the statistical regularities of language, and *functional* linguistic competence, which denotes the ability to *use* language.[79] In practice, it is functional competence that enables humans to accomplish concrete goals. This point is particularly important given the simplistic association of legal work with the production of text. Lawyers do not just "produce text" but *use* language as a medium to convey legal expertise. In contrast, LLMs excell only at formal competence and cannot solve problems *by means of* language. Language use is premised on the prior development of world models that inform how the "physical and social world works."[80] Similarly, the ability to solve problems or devise purpose-driven communications requires knowledge of the causal properties and affordances of everyday objects (aka: common sense!)[81] as well as the ability to model physical and social processes, often referred to as

[77] *PST Energy 7 Shipping LLC & Anor v OW Bunker Malta Ltd & Anor* [2016] UKSC 23 (11 May 2016)

[78] Laura Ruis et al., The Goldilocks of Pragmatic Understanding: Fine-Tuning Strategy Matters for Implicature Resolution by LLMs (2023) 37th Conference on Neural Information Processing Systems (NeurIPS 2023).

[79] Kyle Mahowald et al., 'Dissociating Language and Thought In Large Language Models: a Cognitive Perspective' (2023) https://arxiv.org/abs/2301.06627

[80] Yejin Choi, The Curious Case of Commonsense Intelligence (2022) Daedalus 139, 140; Yann Lecun, A Path Towards Autonomous Machine Intelligence (27 June 2022) https://openreview.net/forum?id=BZ5a1r-kVsf 3; Tyler Millhouse et al., Embodied, Situated, and Grounded Intelligence: Implications for AI' (2022) https://arxiv.org/abs/2210.13589

[81] McCoy et al (n 6) 45

"functional grounding."[82] The latter is, in turn, commonly associated with reasoning[83] and, notwithstanding ongoing progress, it is generally acknowledged that LLMs lack the ability to do so.[84] At present, reasoning in LLMs remains brittle, task-specific and insufficient for the resolution of problems requiring legal analyses.[85] Sophisticated pattern matching can, to an extent, mimic reasoning by replicating word-strings that reflect fixed patterns memorized during training.[86] LLMs cannot, however, reliably perform tasks requiring multi-step or complex reasoning that require the ability to generalize beyond their training data.[87] As is the case with understanding, the ability to use language derives from embodied experiences.[88] The observation that understanding and using language require knowledge of the physical world and shared experiences of *specific* situations[89] is particularly relevant because the meaning of words in a contract often depends on the parties' background knowledge[90] and cannot be disassociated from the context-specific use of language.[91]

It is tempting to conclude this paper here. How could a technology that does not understand the meaning of words be of any use to perform a task requiring the skillful manipulation of words and hair-splitting analyses of individual provisions? After all, the interpretation of contracts often involves years of litigation, with millions of dollars turning on the meaning of a simple word sequence[92] or on the position of a single adjective.[93] Common sense dictates that "something" that does not understand what a contract is (and has never contracted!), could never generate a contract. Yet, as the generated text often looks impressive, I must proceed in order dispel any remaining doubts as to the capabilities of LLMs and fully convey the risks of their deployment.

**Knowing & Applying the Law**

It can be endlessly debated whether contract drafting requires legal expertise. Admittedly, many contractual documents are "prepared" (i.e. copied or downloaded etc.) by non-lawyers and many contracts are created by junior lawyers who absent-mindedly customize templates. Such observations miss the point. The fact that many contracts are created without legal advice does not mean that contract drafting does not require legal knowledge in general. Generating contractual documents that do not fit the laws of the relevant jurisdiction is a waste of computing resources: it neither assists the parties in achieving their goals, nor improves upon the common practice of re-using random documents found online. The purpose of a contractual document is, after all, to ensure enforceability and to provide a modicum of legal certainty. Abstracting from small value exchanges that are executed immediately, contractual documents are usually prepared when it is necessary to address commercial and legal risks or to meet regulatory requirements. Not for the sake of producing words. Even standard form contracts governing low-value consumer transactions must comply with exacting consumer protection regimes. [94] Moreover, contract drafting often necessitates an exercise in pessimism and clairvoyance as many provisions address the consequences of incorrect- or non-performance, as well as unforseen future events. In fact, provisions limiting or excluding liability, indemnities and liquidated damages are the most negotiated terms[95] and those who draft them must understand the risks inherent in a given transaction and know the relevant legal principles, including the manner of estimating the greatest possible loss that could result from the breach or otherwise protecting a party's legitimate interest. Clauses excluding liability or reducing the available remedies, are particularly difficult to draft as they require knowledge of the additional principles applicable to their construction.[96] After all, "the more valuable the right being abandoned or curtailed, the clearer the language will need to be."[97] To generate such clauses, or to comply with industry-specific instruments, LLMs must know the law. To what extent, if any, can they? Of course, references to "knowledge" cannot be taken literally as LLMs only "know" what words typically follow other words.[98] Anthropomorphic terms like "learning" or "knowledge" obfuscate the fact that LLMs replicate word distributions, not facts or expertise. Nonetheless, as the generated text is usually plausible and seemingly correct (at least at a sentence level), LLMs must "know something." To understand the extent of their "legal expertise," it is worth introducing the concept of parametric knowledge. Denoting the information encoded in an LLM's parameters,[99] parametric knowledge can be regarded as a side-effect of the memorization of high-probability word sequences in the training corpora.[100] Pertinently, parametric knowledge does not take the form of explicit facts but that of strengthened connections, or weights, between layers in neural networks.[101] If a word sequence is popular, then the connections associated with words in such sequence are strengthened. For example, if the phrase "contracts are enforceable agreements" frequently occurs in the training corpus, then the LLM will assign a higher probability to 'agreement' if it is preceded by 'contract' and 'enforceable.'[102] "Knowing" that these words often appear together, the LLM may generate the aforementioned string when prompted with "what is a contract?" It will do so because this pattern is *stored* in its weights, not because it *knows* the law. As the training corpora contain the majority of publicly available legal sources, it can be assumed that the parametric knowledge of most LLMs stores a lot of legal information.[103] Nonetheless, four additional observations convey the dangers of relying on parametric "legal knowledge."

[82] Thomas Carta et al, 'Grounding Large Language Models in Interactive Environments with Online Reinforcement Learning' (2023).
[83] Mahowald (n x) 14
[84] Melanie Mitchel et al, Comparing Humans, "GPT-4, and GPT-4V On Abstraction and Reasoning Tasks" (2024) Proceedings of the LLM-CP Workshop, AAAI 2024; https://arxiv.org/abs/2311.09247; Thomas McCoy et al., When a language model is optimized for reasoning, does it still show embers of autoregression? An analysis of OpenAI o1 (4 Oct, 2024) https://arxiv.org/abs/2410.01792.
[85] Parshin Shojaee et al. The illusion of thinking: Evaluating reasoning in puzzle-based frameworks for large language models. arXiv preprint arXiv:2506.06941, 2025; Iñaki Dellibarda Varela, Pablo Romero-Sorozabal, Eduardo Rocon, and Manuel Cebrian. Rethinking the illusion of thinking, 2025; Alex Lawsen et al. The illusion of the illusion of thinking. arXiv preprint arXiv:2506.09250, 2025.
[86] Chris Su and others, 'Limits of Emergent Reasoning of Large Language Models in Agentic Frameworks for Deterministic Games' (arXiv, 12 October 2025) doi:10.48550/arXiv.2510.15974; B. M. Lake, M, Baroni, Human-like systematic generalization through a meta-learning neural network (2023) 623 *Nature* 115–121; Laura Ruis et al., Procedural Knowledge in Pretraining Drives Reasoning in Large Language Models (19 Nov, 2024); Zhaofeng Wu Et al., Reasoning or reciting? Exploring the capabilities and limitations of language models through counterfactual tasks, in Kevin Duh et al., (eds.), Proceedings of the 2024 Conference of the North American Chapter of the Association for Computational Linguistics: Human Language Technologies (Volume 1: Long Papers), pp. 1819–1862.
[87] Subbarao Kambhampati, Can Large Language Models Reason and Plan? (2024) Annals of the New York Academy of Sciences. 10.1111/nyas.15125. 1534:1. (15-18); Marianna Nezhurina et al., Alice in Wonderland: Simple Tasks Showing Complete Reasoning Breakdown in State-Of-the-Art Large Language Models (5 Mar 2025) arXiv:2406.02061v5; Dagmara Panas et al., Can Large Language Models put 2 and 2 together? Probing for Entailed Arithmetical Relationships (??); Lukas Berglund et al., The Reversal Curse: LLMs Trained on "A is B" Fail to Learn "B is A" (21 September, 2023); Freda Shi, et al., 'Large language models can be easily distracted by irrelevant context' in Andreas Krause et al. (eds.), (2023) Vol 202 of Proceedings of Machine Learning Research, pp. 31210–31227.
[88] George Lakoff and Mark Johnson, Philosophy in the Flesh: the Embodied Mind and the Challenge to Western Thought (New York, Basic Books, 1999) 4; Harry M. Collins, "Embedded or embodied? A review of Hubert Dreyfus' what computers still can't do," (1996) 80 Artificial Intelligence 99; Cedegao E. Zhang et al., Grounded Physical Language Understanding with Probabilistic Programs And Simulated Worlds (2023) ICLR 2023.
[89] Morten H. Christiansen, Nick Carter, *The Language Game* (Penguin Random House, 2022) 62; Wittgenstein, L. (1953). Philosophical Investigations. Basil Blackwell, Oxford
[90] Lord Hoffmann in *Chartbrook Ltd v Persimmon Homes Ltd* [2009] UKHL 38, [2009] 1 AC 1101, para 14; *Investors Compensation Scheme Ltd v West Bromwich Building Society* [1998] 1 WLR 896 Lord Hoffmann (pp 912-913)
[91] *Rainy Sky SA v Kookmin Bank* [2011] UKSC 50 at 14-30; *Arnold v Britton* [2015] UKSC 36 at 14-22; *Wood v Capita Insurance Services Limited* [2017] UKSC 24 at 8-15; see also Wittgenstein, at 66; Kaya Stechly et al., Chain of Thoughtlessness: An Analysis of CoT in Planning (8 May 2024); for a more recent evaluation of the latest LLMs, see: Karthik Valmeekam et al., 'LLMs Still Can't Plan; Can LRMs? A Preliminary Evaluation of Openai's 01 on Planbench' (20 September 2024).
[92] *Soteria Insurance Limited (formerly CIS General Insurance Limited) v IBM United Kingdom Limited* [2022] EWCA Civ 440; where the argument came down to the meaning of the word "otherwise" in one provision.
[93] *Cantor Fitzgerald v Yes Bank* [2024] EWCA Civ 695

*First*, the quality of the parametric knowledge depends on the quality of the training corpora.[104] As the latter generally derive from uncurated online sources, which abound in outdated and incorrect information,[105] parametric legal knowledge may also be outdated or incorrect. Crucially, even if the training corpora contained correct legal information, such information would not be memorized if it was less frequent (i.e. lower probability) than incorrect legal information.[106] To be memorized, word sequences must be "high-probability," not "high-quality."[107] Unsurprsingly, LLMs are generally more "knowledgeable" about popular judgements but less "knowledgeble" about judgements from lower courts.[108] Similarly, newer cases, even those that overrule or significantly affect the application of prior precedent, are statistically underrepresented online and thus less likely to be memorized. An LLM's parametric knowledge will, for example, contain more word sequences from *Dunlop Pneumatic Tyre v. New Garage and Motor*[109] than from *Talal El Makdessi v Cavendish Square Holdings BV,*[110] as the former case is older and had been frequently mentioned in both primary and secondary a legal sources for over 100 years. LLMs focus on frequency, not recency and cannot distinguish between legal sources that remain relevant and legal sources that have been rendered obsolete by newer precedent or legislative changes.[111]

[94] In the UK, for example, contracts with consumers must comply with the Consumer Rights Act 2015, Consumer Contracts (Information, Cancellation and Additional Charges) Regulations 2013; The Consumer Protection Unfair Trading Regulations 2008, SI 2008/1277, to name a few.

[95] World Commerce & Contracting Report on the latest Most Negotiated Terms https://www.worldcc.com/Portals/IACCM/Resources/11463_0_Most-Negotiated-Terms-2022.pdf

[96] *Gilbert-Ash (Northern) Limited v Modern Engineering (Bristol) Limited* [1974] AC 689, Lord Diplock said at 717H

[97] *Stocznia Gdynia SA v Gearbulk Holdings Limited* [2010] QB 27, Moore-Bick LJ at X; *BHP Petroleum Limited v British Steel PLC* [2000] 2 Lloyd's Rep 277.

[98] Shanahan, M. (2022). Talking about large language models (2022) arXiv:2212.03551, 5

[99] The term "parameter" refers to the connection between layers in a neural network; Nicholas Carlini et al., 'Quantifying memorization across neural language models,' in: The Eleventh International Conference on Learning Representations, 2023. URL https://openreview.net/forum?id=TatRHT_1cK.

[100] [XXX]; for a broader discussion as to how such "knowledge" is stored, see: Peter Hase et al., 'Does Localization Inform Editing? Surprising Differences in Causality-Based Localization vs. Knowledge Editing in Language Models' (Arxiv October 16, 2023)

[101] Fabio Petroni et al., *Language Models as Knowledge Bases?* PROC. 9th INT'L. JOINT CONF. NLP 2463 (2019).

[102] In neural networks, neurons are interlinked by a series of weighted connections, each of which roughly corresponds to the strength of the relationship between inputs; *see*: Russel & Norvig, (n x) 801-802.

[103] Matthew Dahl, Varun Magesh, Mirac Suzgun and Daniel E Ho, 'Large Legal Fictions: Profiling Legal Hallucinations in Large Language Models' (2024) 16 Journal of Legal Analysis 64–93.

[104] R. Stuart Geiger et al., *"Garbage in, garbage out" revisited: What do machine learning application papers report about human-labeled training data?* 2 QUANTITATIVE SC. STUD. 795 (2021).

[105] Training data predominantly stem from the Common Crawl archive of Internet text scraped from billions of URLs. As a large amount of its content is undesirable for training, LLM builders train their models on filtered samples of Common Crawl, such as Alphabet's "Colossal Clean Crawled Corpus," see: Stefan Baack, 'A Critical Analysis of the Largest Source for Generative AI Training Data: Common Crawl,' in: Proceedings of the 2024 ACM Conference on Fairness, Accountability, and Transparency (FAccT '24). Association for Computing Machinery, New York, NY, USA, 2199–2208.

*Second*, the popular assumption that legal knowledge could emerge from massive amounts of training data[112] is nonsensical.[113] Knowledge cannot be "brute forced" from word distributions,[114] because it cannot be reduced to pattern recognition or the calculation of word probabilities.[115] The acquisition of knowledge requires understanding.[116] As LLMs operate at the level of word patterns, they cannot separate the facts of a case from its *ratio decidendi*,[117] interpret and reconcile inconsistent statutory provisions – not to mention disentangle double negatives or subtle differences in the manner judicial opinions are formulated.[118]

*Third*, legal knowlegde cannot be operationalized without world knowledge**.** Drafting indemnities or exclusion clauses requires legal knowledge but drafting delivery clauses requires knowledge of physical relationships between objects. The same can be said about "synchronising" delivery with the passing of property and the transfer of risk. A lot of world knowledge is, however, tacit and must be acquired through embodied experiences. Training corpora are, however, silent about many fundamental facts about the world because training texts, such as Wikipedia pages, news websites or even legal sources, do not state what is generally known."[If] someone says, 'I bought groceries', he is unlikely to add that he used money to do so, unless the context made this fact surprising or in question."[119] The frequency with which certain facts are written about need not correspond to their likelihood or importance in the real world. The resulting discrepancy between reality and its description in text is known as reporting bias.[120] As unusual events or exceptional attributes of objects are written about more frequently than common ones,[121] LLMs often memorize word sequences referring to atypical facts or opinions but may fail to store common knowledge. At present, the creation of depositories containing basic world knowledge remains an open research question.[122]

---

[106] Inbal Magar, Roy Schwartz, Data Contamination: From Memorization to Exploitation, PROC. 60th ANN. MEETING ASSOC'N COMPUTATIONAL LINGUISTICS 157 (2022); Nikhil Kandpal et al., 'Large language models struggle to learn long-tail knowledge' in: Andreas Krause et al., editors, Proceedings of the 40th International Conference on Machine Learning, volume 202 of Proceedings of Machine Learning Research, pages 15696–15707. PMLR, 23–29 Jul 2023.
[107] Alex Mallen et al., 'When Not to Trust Language Models: Investigating Effectiveness of Parametric and Non-Parametric Memories' (2023) Proceedings of the 61st Annual Meeting of the Association for Computational Linguistics Volume 1: Long Papers, 9802–9822
[108] Dahl et al., (n X)
[109] [1915] A.C. 79
[110] [2013] EWCA 1539
[111] Li Zhang, Jaromir Savelka, Kevin Ashley, 'Do LLMs Truly Understand When a Precedent Is Overruled?' (2025) arXiv preprint arXiv:2510.20941 https://arxiv.org/pdf/2510.20941.pdf
[112] Chris Anderson, 'The End of Theory: The Data Deluge Makes the Scientific Method Obsolete,' Wired June 23, 2008 https://www.wired.com/2008/06/pb-theory/
[113] Subbarao Kambhampati, 'Polanyi's revenge and AI's new romance with tacit knowledge' (2021) 64 Commun. ACM 2, 31–32.
[114] Stefanie Krause and Frieder Stolzenburg. 'From Data to Commonsense Reasoning: The Use of Large Language Models for Explainable AI' (2024). Arxiv 4 July, 2024
[115] Varun Magesh et al., Hallucination-Free? Assessing the Reliability of Leading AI Legal Research Tools (2025) 22 Journal of Empirical Legal Studies 216
116 Luciano Floridi, *The 4th Revolution, How the Infosphere is Reshaping Human Reality* (OUP 2014) 130
[117] Aaron Tucker, Colin Doyle, "If You Give an LLM a Legal Practice Guide" in Proceedings of the 41st International Conference on Machine Learning, Vienna, Austria. PMLR 235, 2024.
[118]
[119] Jonathan Gordon and Benjamin Van Durme, 'Reporting Bias and Knowledge Acquisition' (2013) Proceedings of the Workshop on Automated Knowledge Base Construction 25-30
[120] Id.
[121] Id.
[122] For example, the CYC project, which represents several hundred person-years of expert human labor, encodes commonsense knowledge in a quasi-logical symbolic system and enables automated inference to do commonsense reasoning, *see*: Douglas B. Lenat et al., *CYC: Towards Programs with Common Sense*, 33 COMM. ACM. 30 (1990).

Lastly, once training is completed, parametric knowledge is “frozen in time.” When genertaing text, LLMs lack access to current information, including changes in legal instruments or recently decided cases. Parametric knowledge is not only inherently incomplete and often incorrect, but also outdated.

## Applying the Law

Contract drafting requires the ability to reason about the law in the context of specific transactions. Laws may imply terms, prescribe terms or require that terms be fair or reasonable. Even if, for the sake of argument, an LLM’s parametric knowledge contained all legal sources in their correct and current form, the model would still be unable to utilize them. For example, prompted to generate a consumer contract in England, an LLM must know (understand?) that consumer contracts must be fair in the sense prescribed by the Consumer Rights Act 2015.[123] An LLM can, in principle, “know” the text of this instrument but it cannot generate provisions complying with its substance. Lacking any comprehension of “fairness,” LLMs cannot consider the contractual subject matter or refer to “all the circumstances existing when the term was agreed.”[124] Knowing how “fair” relates to other words in the training corpora is unrelated to the question whether a particular term is fair, especially given that a “fair term” need not contain the word itself. Consequently, LLMs cannot generate exclusion clauses that are fair in light of the indicative list of potentially unfair terms listed in the Act.[125] They cannot extrapolate from the provided examples because they “see” word patterns, not principles. The same observations apply to a myriad of other legal instruments, such as the Unfair Contract Terms Act 1977 or the Contracts (Rights of Third Parties) Act 1999.[126] Knowing how “privity” relates to other words is unrelated to understanding the concept of privity or enforceability. The problem is aggravated in more complex transactions, which often require the reconciliation of several legal instruments.[127] An indicated, the reasoning capabilities of LLMs are generally insufficient to deal with the complexities of legal language.[128]

Some readers might observe that the training corpora already comprise thousands of contracts that comply with the relevant laws. Consequently, even if LLMs only recombined provisions common in such documents, the generated contracts would be legally compliant. LLMs do not, however, recombine the relevant and legally compliant provisions but only replicate word patterns frquently occurring in such provisions. They never cross the semantic threshold[129] that would allow them to evaluate which patterns to replicate or which contractual provisions to include when generating a specific document. A generated contract for the sale of a Tesla car will derive from *all* high-probability word sequences in the training corpora that contain (or co-occur with) the words “sale,” “car” and “Tesla,” irrespective of their quality or adequacy. Some of those word sequences may even concern the sale of Tesla shares, not cars; others may derive from prospectuses or marketing materials. Moreover, the generated “word recombinations”may not suit the transaction at hand. LLMs cannot critically evaluate which words should be recombined for a *particular* contract or differentiate between contracts from dissimilar jurisdictions. Provisions that work in the context of one contract may be detrimental in another. Provisions that are acceptable (and common!) in consumer contracts in the United States may be unacceptable in the European Union.[130] Training corpora are not organized along jurisdictional lines and LLMs do not distinguish between the laws of Scotland and the laws of Australia. It must also be remembered that when new legal instruments come into force or when new cases are decided, then

[123] Consumer Rights Act 2015, Section 62 (4) and (5)
[124] Consumer Rights Act 2015, Section 62 (4) and (5) (a), (b)
[125] Consumer Rights Act 2015, Section 63
[126] some instruments prescribe that certain contracts contain specific provisions, for example, the UK Housing Grants, Construction and Regeneration Act 1996 puts construction contracts within a tight regulatory framework that prescribes certain provisions, such as many aspects of payment.
[127] Prompted to generate a contract for the carriage of goods by sea, an LLM would not only have to understand “distance” or “sea” but also reconcile several competing regimes, such as the Hague Visby Rules of 1968, the Hamburg Rules, the Rotterdam Rules of 2009 and the Special Drawing Right (SDR) of the International Monetary Fund.
[128] Li Zhang, Matthias Grabmair, Morgan Gray and Kevin D Ashley, ‘Thinking Longer, Not Always Smarter: Evaluating LLM Capabilities in Hierarchical Legal Reasoning’ (9 October 2025) arXiv:2510.08710; Jonathan H Choi, 'Off-the-Shelf Large Language Models Are Unreliable Judges' (SSRN, 8 April 2025, last revised 7 May 2025) https://ssrn.com/abstract=5188865 accessed 9 December 2025;
[129] Floridi (n x) 135
[130] See, for example, the rampant use of arbitration clauses in consumer contracts in the US, *AT&T Mobility LLC v. Concepcion*, 563 U.S. 333 (2011)

contracts drafted with such developments in mind will take time to proliferate through the training corpora in sufficient quantities to become memorized in a model's parametric knowledge. We have, for example, only 13 years' worth of liquidated damages provisions drafted on the basis of *Talal El Makdessi*[131] and over 110 years of provisions relying on *Dunlop Pneumatic Tyre Company.*[132] In sum, even if LLMs were trained on millions of legally compliant contracts, there would be no guarantee that the generated contracts would also be compliant. Text generation is driven by probability, not by substantive relevance or regard to legal considerations.

## Potential Solutions

There are several methods of adapting LLMs to downstream tasks and alleviating their inability to reason as well as the shortcomings of parametric knowledge. None of them can, however, significantly improve the chances of generating viable contracts. Anticipating vigorous objections that I am underplaying the possibility of improving the performance of LLMs through training or at inference, it is worth sketching some of the relevant techniques.

### Training on Legal Texts?

Training LLMs on legal text would not enhance their ability to generate contracts.[133] An LLM trained on all legal sources from all common law jurisdictions, would "only" learn word distributions specific to such sources. Similarly, LLMs trained on billions of contracts would learn word distributions specific to such contracts and memorise word sequences frequently occuring in such documents, such as "notwithstanding the foregoing," "herewith agree" or "indirect or consequential damages." LLMs would, however, remain incapable of selecting which documents or word sequences should be included in a particular generation task. Even if, for the sake of argument, there existed training corpus comprised of high-quality contractual documents reviewed by a battalion of experienced counsel, it would remain impossible to ensure that the contracts generated on the basis of such "training contracts" retained their quality. While improving the quality and relevance of the training corpora represents a commonsensical approach, it does not counterbalance the fundamental characteristics of LLMs. improving the text of the training corpora does not change the fact that LLMs can only replicate high-probability word sequences derived from such corpora and remain isolated from the context in which a document is generated. The octopus will always stay in the water, no matter what A and B are talking about.

An aside: why would anyone generate a contract representing an anodyne, probabilistic assemblage of word sequences from thousands of " high-quality contracts" instead of using one of those "high-quality contracts" in their original form?[134]

Additional challenges arise during post-training, which aims to adapt LLMs to specific tasks. While post-training requires less training data than pre-training, it requires a significant amount of human expertise as the relevant techniques rely on labelled input-output examples[135] and human evaluations of the generated outputs.[136] Creating such examples and performing the relevant evaluations is challenging given the indispensable expertise as well as the potential for error and disagreement. It is difficult to label contracts and contractual provisions as "good" or "bad" in isolation, without regard to the transactions they govern. Even if significant resources were committed to preparing examples of provisions of varying quality, as well as instructions as to how to evaluate them, there will always be subjective and context-dependent factors that may render such efforts pointless. There is no perfect contract of sale that can accommodate all transactions in goods or a "universally applicable" indemnity that can be inserted into every contract.[137] To re-state the obvious: the adequacy of a contract or contractual provision cannot be evaluated in isolation from the context of a transaction.

[131] [2015] UKSC 67
[132] [1914] UKHL 1
[133] E Martinez, F Mollica, E Gibson, 'So much for plain language: An analysis of the accessibility of United States Federal Laws over time' Available at SSRN: https://ssrn.com/abstract=4036863 (2022).
[134] (assumedly, with some customizations)
[135] Radford et al., (n x) 2, 3; Brown et al., (n x) 6.
[136] Long Ouyang et al., *Training Language Models to Follow Instructions with Human Feedback*, 36 PROC. CONF. ON NEURAL INFO. PROCESSING SYS. (2022).

In sum, LLMs trained on contractual documents would be more proficient at *mimicking* legal language. The generated "contractual deepfakes" would appear even more authoritative, especially to non-lawyers. They would, however, remain detached from the substance of the law and the circumstances of the transaction. No amount of training data or curated examples can convey the full transactional context and the applicable legal framework.

Sophisticated Prompting

Technical papers abound in descriptions of prompting techniques that are supposed to adapt LLMs to concrete tasks.[138] As contracts are drafted for specific transactions, prompts could include examples of suitable provisions, the applicable legal principles (e.g. "liquidated damages must not be deterrent") as well as the commercial details, including transcripts of negotiations or descriptions of the transactional background. In theory, carefully crafted prompts *could* lead to viable contractual documents or, at least, increase the probability of their generation. In practice, however, all prompting techniques encounter similar obstacles.

*First*, prompting requires not only syntactic rigor, but also iterative evaluations of each output generated in response to specific prompt formulations.[139] Given the superficial plausibility of the generated text and the difficulty of detecting hallucinations in longer passages, repetitive reviews are time-consuming. As indicated, LLMs reduce the time to generate text – not the time required to read text. To aggravate matters, LLMs are non-deterministic and thus inherently unpredictable. Identically worded prompts may thus yield different outputs[140] and even meticulously crafted prompts cannot ensure that LLM will generate consistently relevant text.[141] Is it worth expanding resources on engineering a "perfect prompt" if its output(s) may not be reproducible?

*Second*, LLMs are prompt-sensitive.[142] Instructions of equivalent substance expressed in slightly different words will yield different outputs.[143] The addition of a single word may significantly affect the generated text, even if such addition should be inconsequential. [144] Crucially, LLMs are generally unable to differentiate between relevant and irrelevant details or distinguish instructions concerning the type of contract from information about the transaction. [145] Moreover, being sensitive to high-probability words, models generally perform better when prompts refer to facts or concepts frequently mentioned in their training corpora.[146] This means that even if a prompt conveys sufficient information about the context of a transaction, such information may be ignored and the model may fall back on regurgitating common contractual language..

*Third*, many LLMs feature larger context windows (i.e. the amount of text that can be inserted into the prompt)[147] and can thus be provided with more information. Larger context windows are only beneficial, however, if such information can be fully and correctly utilized, a metric known as "recall

[137] If such perfect contracts of sale or perfect indemnities were in existence, there would be no need to generate new ones in the first place.
[138] Z Lin, 'How to write effective prompts for large language models' (2024) 8 Nat Hum Behav 611, 612.
[139] Kaiyan Chang et al., Efficient Prompting Methods for Large Language Models: A Survey (1 April, 2024) https://arxiv.org/abs/2404.01077
[140] Shuyin Ouyang et al., LLM is Like a Box of Chocolates: the Non-determinism of ChatGPT in Code Generation (5 August, 2023); Luciano Floridi, Massimo Chiriatti, *GPT-3: Its Nature, Scope, Limits, and Consequences*, 30 MINDS & MACH. 681, 687 (2020).
[141] An LLMs probabilistic behaviour may, for example, be affected by temperature settings, which controls the randomness of the model's output. A higher temperature (e.g., 1.0) increases response diversity, while a lower temperature (e.g., 0.1) makes models more deterministic.
[142] Z. Zhao et al., 2021. Calibrate before use: Improving few-shot performance of language models. In ICML 12697–12706. PMLR.
[143] J.D. Zamfirescu-Pereira, 'Why Johnny can't prompt: How Non-AI Experts Try (and Fail) to Design LLM Prompts (2023) CHI'23 April 23-28; Adam Roegiest et al., A Search for Prompts: Generating Structured Answers from Contracts (2023) https://arxiv.org/abs/2310.10141; Noam Kolt, 'Predicting Consumer Contracts' (2021) 37 Berkeley Tech L J 71.
[144] F. Shi, et al., 2023a. Large language models can be easily distracted by irrelevant context. In ICML, 31210–31227. PMLR.
[145] A. Webson and E Pavlick, 'Do Prompt-Based Models Really Understand the Meaning of Their Prompts?' In Proceedings of the 2022 Conference of the North American Chapter of the Association for Computational Linguistics: Human Language Technologies, 2300–2344.
[146] McCoy et al (n 6) 29, 30.
[147] For example, Google's Gemini 1.5 has a context length that enables it to process more than ten times the entirety of the 1440 page book (or 587,287 words) "War and Peace;" see: Gemini 1.5: Unlocking multimodal understanding across millions of tokens of context (2024) Google DeepMind

performance."[148] As LLMs tend to ignore text provided in the middle-portions of the prompt[149] and often struggle to process longer passages, providing detailed information about the contractual background or the applicable legal principles may be futile. LLMs may "overlook" instructions concerning the choice of law, the payment terms or other important legal details. Recall performance is also degraded when prompts contain too much information[150] or when such information conflicts with the model's parametric knowledge, that is, when the specific information in the prompt differs from the more general (and hence more frequent) information in the training corpora.[151] As the exact "contents" of parametric knowledge are generally unknown, such conflicts cannot be avoided. Providing detailed instructions may thus, counterintuitively, fail to yield better outputs.[152] Moreover, larger amounts of information may not improve performance in the case of open-ended questions as well as difficult or novel tasks.[153]

### Retrieval Augmented Generation?

The shortcomings of parametric knowledge could be improved by a technique known as Retrieval Augmented Generation, or "RAG,"[154] which involves the provision of information from external sources. Unlike parametric knowledge, external sources can be frequently updated and adjusted to specific tasks when the LLM is used. When, for example, prompted to generate a consumer contract, the LLM could draw from curated sources containing the latest developments in consumer law or examples of legally compliant contractual provisions. In practice, the effectiveness of RAG depends on the quality of such sources and on the ability to provide models with legal knowledge in a distilled, easily processable form. As LLMs can neither understand nor reason about legal texts, they are generally unable to distill the principles from legal sources in order to apply them to actual scenarios.[155] Similarly, connecting LLMs to repositories of contractual provisions or templates may be ineffective. How would LLMs "know" which provisions are suited for a given transaction? How would they select the optimal template given their inability to understand the context in which a contract is generated? Additional challenges concern the retrieval process itself, as LLM cannot distinguish between relevant and irrelevant information[156] or resolve conflicts between conflicting sources. [157]

In sum, neither prompting nor RAG can change the fact that LLMs are unable to reason or to apply legal knowledge to specific transactions.

## Connecting Some Dots

Equipped with a better understanding of their technical attributes, we can revisit the question whether LLMs can generate contracts. To appease technology enthusiasts, I acknowledge that LLMs can generate documents that *look* like contracts and *may* be adopted by the parties. Parties enter into contracts on their own terms and can sign any contract – or contractual deepfake - they want.[158] Once the document is signed, they are bound.[159] In principle, it is irrelevant whether its provisions were generated by an LLM or drafted by a lawyer.[160] As indicated, the objective theory of contract disregards the technical provenance of the text. Nothing in existing caselaw suggests that such "artificial" should affect the legal status of the document or subject it to different rules. It seems equally irrelevant that the generated contract contains terms that are unfair or unfavorable to one or even both parties.[161] This approach reflects the primacy of the written document[162] and the importance of commercial certainty, as embodied in the objective theory of contract.[163] It is trite law that once the parties have to all outward appearances agreed in the same terms on the same subject-matter,[164] then neither can, generally, invoke some unexpressed qualification or reservation to show that they had not agreed to the terms to which they had appeared to agree.[165] If, in other words, the parties decide to adopt a generated document in its "original" form, without extensively reviewing its contents, they bear the resulting risks. Yet, can it really be said that the generated document is a contract, just because it *mimics* a contract and contains provisions commonly encountered in contracts? Does it become a contract because the parties decided to sign it? As indicated, the adoption of a particular document by the parties need not be an indicator of its quality or even suitability. It is, after all, statistically improbable that the generated contractual document is viable from the outset, at least if viability is associated with legal compliance *and* suitability to govern a particular transaction. Instead of debating whether LLMs can generate viable contracts two separate questions are apposite: (a) what are the risks of deploying LLMs for this task? and (b) is it more efficient to generate contracts or to draft them in a traditional manner? The first question highlights the dangers of indiscriminately signing generated documents and the difficulties of "repairing" them with the available legal tools; the second considers the resources required to review and amend generated documents as compared to the common practice of customizing templates. Indirectly, the first question concerns the costs of litigating generated contracts after they are signed, while the second concerns the costs of reviewing such documents before they are signed. Answered together, they provide a better idea as to whether LLMs *should* be used to generate contracts.

[148] Traditionally, the concept of "recall" serves as a metric for Information Retrieval (IR) systems, assessing their ability to retrieve relevant information from a corpus given a search query. In LLM evaluation, "recall" is a metric used to evaluate a model's ability to retrieve a factoid from its prompt.

[149] Nelson F. Liu et al., 'Lost in the Middle: How Language Models Use Long Contexts' (July 2023) arxiv.

[150] Providing too much information in the context window, or context stuffing, generally leads to a phenomenon called "context rot," the progressive, non-uniform decline in LLM reliability, see: at https://research.trychroma.com/context-rot

[151] Daniel Machlab, Rick Battle, 'LLM In-Context Recall is Prompt Dependent' (April 14. 2024) Arxiv

[152] Richard Heersmink et al., A phenomenology and epistemology of large language models: transparency, trust, and trustworthiness (2024) 26 Ethics and Information Technology 41

[153] Xiang Li et al., 'Why does in-context learning fail sometimes? Evaluating in-context learning on open and closed questions' (2 July 2024) Arxiv at…

[154] Patrick Lewis, et al., *Retrieval-augmented generation for knowledge-intensive NLP tasks,* ADVANCES IN NEURAL INF PROCESSING SYST 33 (2020), 9459–9474; Kelvin Guu, et al.,. *Retrieval augmented language model pre-training*, INT'L CONF ON MACHINE LEARNING 3929 (2020); G Mialon at al., *Augmented Language Models: a Survey*, ARXIV (Feb. 21, 2023) https://arxiv.org/pdf/2302.07842.pdf.

[155] Lucia Zheng and others, 'A Reasoning-Focused Legal Retrieval Benchmark' (CS&Law 2025, 4th ACM Symposium on Computer Science and Law, Munich, 25–27 March 2025) https://doi.org/10.1145/3709025.3712219; Aaron Tucker, Colin Doyle, "If You Give an LLM a Legal Practice Guide" in Proceedings of the 41 st International Conference on Machine Learning, Vienna, Austria. PMLR 235, 2024; J Chen et al., Benchmarking Large Language Models in Retrieval-Augmented Generation. Proceedings of the AAAI Conference on Artificial Intelligence, 38(16):17754–17762, March 2024. ISSN 2374-3468, 2159-5399.

[156] (Magesh) Amiraz, C., Cuconasu, F., Filice, S., and Karnin, Z. (2025). The Distracting Effect: Understanding Irrelevant Passages in RAG. In *Proceedings of the 63rd Annual Meeting of the Association for Computational Linguistics (Volume 1: Long Papers)*, pages 18228–18258.

[157] Youna Kim and others, 'Reliability Across Parametric and External Knowledge: Understanding Knowledge Handling in LLMs' (arXiv, 19 February 2025) arXiv:2502.13648.

## The Risks of Generating Contracts

Let us assume a generated "contract" was signed without review. When a dispute arises, the parties (finally!) examine the document and realize that "something went wrong" with its wording. While we cannot explore the full spectrum of possible "misgenerations," we can consider several situations depending on whether the generated document is enforceable (or not!) or whether it is enforceable but not viable or even commercially nonsensical. We must also consider the possibility of "repairing" the generated document with any of the mechanisms available in contract law. We could thus speak of a sliding scale of "bad generation," ranging from those "contracts" that are patently nonsensical to those that appear viable but may lead to undesirable results for one or both parties, if enforced in accordance with their wording. Alternatively, we could also contemplate the number of repairs necessary to make the contract work in practice, leaving aside the question whether the final result would be desirable for both parties. Needless to say, each situation must be evaluated in light of the objective theory of contract, which focuses on the outward manifestations of agreement without regard to the fact that one or both parties have not read or understood the document.[166] As observed in one case, "to make the question of consensus, in a case based on a document, depend on the intellectual or linguistic capacity of a party would cause great uncertainty which an objective analysis is intended to overcome."[167] Similarly, great uncertainty would result if the either party was allowed to raise objections on the basis that the text of the contract was generated or resulted in disadvantageous commercial consequences.

---

[158] *Prime Sight Ltd v Lavarello* [2013] UKPC 22*Printing and Numerical Registering Co v Sampson* [1875], at [466]; *Photo Production Ltd v Securicor Transport Ltd* [1989]

[159] *L'Estrange v F Graucob Ltd* [1934] 2 KB 394

[160] Although, when striking a balance between the indications of the language and the implications of competing constructions the court must consider the quality of drafting, see: Rainy Sky, Lord Mance [26]

[161] S. Patrick Atiyah, 'Freedom of Contract and the New Right' in Patrick, Atiyah *Essays on Contract* Oxford, Clarendon Press 1986, p 355; S. Patrick Atiyah, *The Rise and Fall of Freedom of Contract* Oxford, Clarendon Press 1979, p 227.

[162] *FSHC Group Holdings Ltd v Glas Trust Corporation Ltd* [2019] EWCA Civ 1361 at 173

[163] Michael Furmston, *Cheshire, Fifoot & Furmston's Law of Contract* (16th edn OUP 2012) 41; Hugh Beale ed, *Chitty on Contracts* (35th edn, Sweet & Maxwell 2024) paras 1-027 to 1-0291; *Maple Leaf Macro Volatility Master Fund v Rouvroy* [2009] EWHC 257 (Comm) [223], [224]

[164] See Chitty 1-053 citing *Falck v Williams [1900] A.C. 176*; *Pagnan SpA v Fenal Products Ltd [1987] 2 Lloyd's Rep. 601* at 610; *Guernsey v Jacob UK Ltd [2011] EWHC 918 (TCC), [2001] 1 All E.R. (Comm) 175* at [41]; *Global 5000 Ltd v Wadhawan [2011] EWHC 853 (Comm), [2011] 2 All E.R. (Comm) 190* at [45]; *VTB Capital Plc v Nutritek International Corp [2013] UKSC 5, [2013] 1 All E.R. 1296* at [140].

[165] See Chitty 1-053 citing *Thoresen Car Ferries Ltd v Weymouth Portland BC [1977] 2 Lloyd's Rep. 614*; *Maple Leaf Volatility Master Fund v Rouvroy [2009] EWCA Civ 1334, [2010] 2 All E.R. (Comm) 788* at [10]; *Air Studios (Lyndhurst) Ltd v Lombard North Central Plc [2012] EWHC 3162 (QB), [2013] 1 Lloyd's Rep. 63* at [5];

In the first situation, the generated "contract" contains so many hallucinated, nonsensical or inconsistent provisions that it cannot be analysed in terms of enforceability. Being nothing but a regurgitation of popular contractual language, it does not describe any obligations in a manner that would enable their performance. The document is objectively useless and cannot be called a "contract." The parties are left with no contractual document and must prove their agreement (if any) by other means. In the second situation, the generated contract is *prima facie* enforceable but does not reflect the parties' agreement. To phrase it differently, from the perspective of the objective theory of contract, the generated document *is* a contract and the parties are bound. Nonetheless, despite being enforceable, the contract is not viable because it does not support the parties' specific transaction or is commercially disadvantageous to one or both parties. In the first situation, the parties end up without a written contract, in the second - they find themselves bound by a contract neither of them wanted. Anticipating the question "how could an LLM generate an enforceable contract?" a clarification is apposite. As the training corpora contain millions of contracts, it is *theoretically* possible for the generated document to contain the minimum substance required for a certain type of transaction. Moreover, even if the generated document contains "gaps," the requirement that contracts be certain and complete[168] is not absolute. Courts are reluctant to deny the existence of an agreement on the grounds of uncertainty or incompleteness if it is clear that the parties intended to be bound or thought that they concluded a contract, especially if they acted upon it.[169] Some degree of imprecision is tolerated as "laymen unassisted by persons with a legal training are not always accustomed to use words or phrases with a precise or definite meaning"[170] and businessman often express themselves "in crude and summary fashion."[171] Incompleteness can also be remedied by implication, where courts resort to business necessity[172] to "make the contract work." Enforceability does not require an exhaustive description of all respective rights and obligations as contract law has ample mechanisms to "fill the gaps." Theoretically, in the second situation, the generated contract *could* meet the minimum requirements of enforceablity and constitute a *generic* contract of the *type* envisaged by the parties. Still, the contract may not be viable in the sense of addressing their needs in light of a *specific* transaction. Even contracts of sale come in hundreds of variations, their provisions differing greatly depending on the thing being sold. Delivery terms suitable for the sale of 100 tonns of coconuts are unsuitable for the sale of an underwater cable. Parties who recklessly sign the generated contract of sale may find themselves bound by a contract that does not reflect their agreement, capture their "real deal"[173] or, worse yet, does not suit their deal at all. Admittedly, such situation is not uncommon in contracts drafted in a traditional manner, particularly in the case of standard form contracts. Parties generally do not read such documents and give little thought to their substance. These commonsensical observations confirm the difficulty of answering the question whether LLMs can generate contracts. They also confirm the necessity to distinguish between the ability to generate generic contracts of a particular type and the ability to generate contracts customized to the parties' idiosyncratic needs. They also point to a broader question: what is the point of generating contracts that do not serve the parties' goals or reflect their agreement? If the parties are indifferent to the contents of their contract, they could use an *existing* generic template instead of wasting resources on generating yet another generic document! In practice, the parties' actions will depend on the degree of discrepancy between their actual agreement and the generated document as well as on the resources (and motivation!) available to the parties to align the document with their transactional goals by means of, for example, rectification or implication.

---

[166] L'Estrange v Graucob [1934] 2 KB 394; *Coys of Kensington Automobiles Ltd v Pugliese [2011]* EWHC 655 (QB), [2011] 2 All E.R. (Comm) 664

[167] *Coys of Kensington Automobiles Ltd v Pugliese [2011]* EWHC 655 (QB), [2011] 2 All E.R. (Comm) 664 at [40]

[168] *May and Butcher Ltd v R* [1934] 2 KB 17 n, HL per Viscount Dunedin at 21; *Carlill v Carbolic Smoke Ball Co* [1893] 1 QB 256 at 262 (CA).

[169] *RTS Flexible Systems Ltd v. Molkerei Alois Muller GmbH* [2010] UKSC 14, [2010] 1 WLR 753

[170] *Scammell v. Ouston* [1941] AC 251

[171] *Hillas v Arcos* (1932) 147 LT 503

[172] *Marks & Spencer Plc v BNP Paribas Securities Services Trust Co 6Jersey) Ltd* [2015] UKSC 72

[173] Macaulay 79

The problems do not end here. In the third situation, the generated contract seems objectively plausible and enforceable, but a thorough review reveals inconsistencies, ambiguities and gaps. While in the second situation the contract is (or can be easily made) enforceable, the third situation is more complex because it is unclear whether such contract *can* or *should* be repaired to make it enforceable. The difference concerns the number of repairs required. As every ill-drafted contract is ill-drafted "in its own way,"[174] the feasibility of its repair depends (again!) on the degree to which the wording has gone wrong. On one hand, courts will strive to preserve rather than destroy bargains,[175] even if this task is not easy.[176] After all, inconsistencies between different parts of a document "are the everyday stuff of contract and of commerce."[177] Assuming that the principles applying to badly drafted contracts also apply to "badly generated" contracts then, if it is clear that something went wrong with the language, there is "[no] limit to the amount of red ink or verbal rearrangement or correction which the court is allowed."[178] On the other, such limit may exist because it is for the parties, not for the courts to make contracts. Consequently, everything depends on the parties' willingness to litigate the generated "contract" and on the courts' ability to repair the document with any of the mechanisms designed to remedy drafting deficiencies. For example, provisions that contravene the law or contain nonsensical wording can be severed.[179] The risk of hallucinations can thus be alleviated by the blue pencil rule, which permits courts to strike out provisions that are void or otherwise unenforceable. While the invalidity of a single provision need not affect the enforceability of the entire contract, it is assumed that such provision is not essential and that the contract could stand without it.[180] It is also tacitly assumed that only *one* provision (or part thereof) needs to be severed or blue-pencilled. The legal position may, however, differ if multiple provisions are inadequate or contravene the law. If the generated contract contains gaps and inconsistencies, those can be remedied by means of rectification and implication.[181] As rectification displaces the natural presumption that the document *is* the agreement,[182] the parties must prove what they have agreed.[183] Courts may also be reluctant to recify multiple "drafting errors" given that rectification is an equitable remedy designed to correct the record of the contract, not to "make" or alter the contract. [184] Similarly, while courts may be inclined to imply one or two terms to give effect to the parties' intention, they may be unwilling to fill in multiple gaps and, again, "make" the contract for the parties. If, then, a large number of generated provisions must be struck out or rectified and if the gaps in the contract cannot be filled with implied terms, the parties might find themselves in the first situation: with no document recording their agreement.

A similar result may ensue if the generated language is unclear and parties resort to "corrective interpretation."[185] It is trite law that corrections and "verbal rearrangements" can only be made if it can be determined what a reasonable person would have understood the parties to have meant. [186] It must, in other words, be clear what correction should be made.[187] Following the motto "that is certain which can be rendered certain (*id certum est quod certum reddi potest*),"[188] courts will aim to give effect to the parties' objective intention, focusing on the contract's substance, not its form.[189] Again, such substance must point to an ascertainable intention.[190] Drafting errors can be remedied by interpretation, if the commercial purpose of the agreement is clear and the document enables a commercially sensible interpretation.[191] Consequently, irrespective of whether we focus on ascertaining the parties' objective intention or on determining the contract's commercial purpose, the generated language must enable the court to make the contract work in practice. If, however, the language is so obscure, contradictory or imprecise that it becomes impossible to determine the parties' intention,[192] we find ourselves (again!) in the first situation. Despite its superficial plausibility, the "contract" cannot stand.

[174] *Arnold v Britton,* Lord Carnwath at [108]
[175] *Scammell v Dicker* 31; *Anglo Continental Educational Group (GB) Ltd v Capital Homes (Southern) Ltd* [2009] EWCA Civ 218, [2009] CP Rep 30, para 13, Arden LJ; see also: *Great Estates Group Ltd v Digby* [2011] EWCA Civ 1120, [2012] 2 All ER (Comm) 361, Toulson LJ
[176] *Astor Management AG v Atalya Mining plc* [2017] EWHC 425 (Comm); [2018] 1 All ER (Comm) 547, [64] Leggatt J
[177] *Scammell v Dicker* 31
[178] *Chartbrook Ltd v Persimmon Homes Ltd* [2009]1 AC 1101 Lord Hoffmann at [25]
[179] *Nordenfelt v Maxim Nordenfelt Guns and Ammunition Co Ltd* [1894] AC 535; *Rose & Frank Co v JR Crompton & Bros Ltd* [1924] UKHL 2, [1925] AC 445.
[180] *Tillman v Egon Zehnder Ltd* [2019] UKSC 32 at 87
[181] *Codelfa Construction Pty Ltd v State Rail Authority of New South Wales* (1982) 149 CLR 337 at 346 Mason J
[182] *FSHC Group Holdings Ltd v Glas Trust Corporation Ltd* [2019] EWCA Civ 1361 at [46], [174]; see also: see Paul Davies, "Rectification versus Interpretation: The Nature and Scope of the Equitable Jurisdiction" (2016) 75 CLJ 62, 78.
[183] *Joscelyne v Nissen* [1970] 2 QB 86; *FSHC Group Holdings Ltd v Glas Trust Corporation Ltd* [2019] EWCA Civ 1361 at 73.
[184] *FSHC Group Holdings Ltd v Glas Trust Corporation Ltd* [2019] EWCA Civ 1361; *Agip SpA v Navigazione Alta Italia SpA ("The Nai Genova")* [1984] 1 Lloyd's Rep 353 at 359; *Oceanbulk Shipping & Trading SA v TMT Asia Ltd* [2010] UKSC 44 at [45] Lord Clarke
[185] *Cherry Tree Investments Ltd v Landmain Ltd* [2012] EWCA Civ 736; [2013] Ch 305, [62] Arden LJ.
[186] *Rainy Sky SA v Kookmin Bank* [2011] UKSC 50
[187] *Chartbrook* [2009] 1 AC 1101, Lord Hoffmann paras 22-24
[188] *Scammell v Dicker* 39
[189] *Arnold v Britton,* Lord Carnwath at [?]
[190] *Scammell v. Ouston* [1941] AC 251, 268 Lord Wright

Counterintuitively, greater problems may arise in situations, where the generated provisions seem objectively clear and unambiguous but, when read together, flout business common sense[193] or lead to nonsensical commercial results.[194] To recall, LLMs can produce seemingly coherent legal language (including sophisticated legalese) but lack world knowledge and common sense. They are also prone to hallucinate. Nonetheless, it is assumed that the parties control the contents of their contracts and, at least theoretically, chose their words deliberately. [195] If, then, the parties have adopted the generated document as theirs and if its language is objectively clear, then it may be difficult to justify departures therefrom.[196] This, in turn, may lead to the unfortunate result of having to enforce a contract that is commercially nonsensical or disadvantageous to one or both parties. The need to avoid nonsensical results[197] does not, after all, provide a *carte blanche* to remedy all "drafting infelicities" through interpretation.[198] Courts cannot rewrite contracts to make them "conform to business common sense,"[199] or "attribute to the parties an intention which they plainly could not have had."[200] Similarly, courts generally do not rewrite contracts to relieve the parties from bad bargains [201] or from the consequences of their imprudence.[202] While bad bargains may be difficult to distinguish from nonsensical bargains, it is indisputable that signing a generated "contract" without extensive review exemplifies such imprudence. In the best-case scenario, the generated and "repaired" contract is enforceable but commercially disadvantageous to one party; in the worst case – it is commercially nonsensical or disadvantageous to both. In the latter situation, the parties would be better off if the contract was unenforceable altogether.

## To Generate or to Customize?

When deciding whether to generate or whether to draft a contract in a traditional manner, we must consider whether using LLMs improves upon existing practices. In principle, drafting rarely starts with a blank page as it is common to tailor the contents of existing contracts to the needs of new transactions or, particularly in law firms, to customize contract templates. In practice, these scenarios are similar as most templates derive from previously used contractual documents. The main differences between them concern the source of the original document and the number of amendments required. The trade-off is always between efficiency in terms of resource requirements and the minimum acceptable quality. It must also be remembered that many industries, such as shipping and construction, already abound in contract templates that provide a widely accepted legal framework.[203] In such instance, there is little benefit in creating new contractual documents.

---

[191] See, e.g. *Homburg Houtimport BV v Agrosin Private Ltd (The Starsin)* [2004] 1 AC 715 where the mistake was a clear omission of words in a standard clause; *Chartbrook Ltd v Persimmon Homes Ltd* [2009] 1 AC 1101, where the transactional background and the internal context of the contract revealed a linguistic mistake in a definition; *Aberdeen City Council v Stewart Milne Group Ltd* [2011] UKSC 56, 2012 SCLR 114 Lord Hope [18] [19] and [22]. (on 'correctional interpretation'); *Pink Floyd Music Ltd and another v EMI Records Ltd.* [2010] EWCA Civ 1429 at 22, citing *City Alliance Ltd v Oxford Forecasting Services Ltd* [2001] 1 All ER Comm 233, Chadwick LJ para 13

[192] *Scammell v. Ouston* [1941] AC 251, 268 Lord Wright

[193] *Antaios Cia Naviera SA v Salen Rederierna AB (The Antaios)* [1985] AC 191, Lord Diplock at 201; Lord Reid in *Wickman Machine Tools Sales Ltd v L Schuler AG* [1974] AC 235, 251

[194] *Arnold v Britton,* Lord Carnwath at [108]

[195] *Arnold v Britton*, L Neuberger 17

[196] *Arnold v Britton*, Lord Neuberger at [18]

[197] *Arnold v Britton,* Lord Carnwath at [108]

[198] *Arnold v Britton*, Lord Neuberger at [18]; Aberdeen, Lord Hope at [18]

[199] Hoffmann LJ (*Co-operative Wholesale Society Ltd v National Westminster Bank plc* [1995] 1 EGLR 97, 99)

[200] *Investors Compensation Scheme Ltd v West Bromwich Building Society* [1998] 1 WLR 896, 912-913, Lord Hoffmann's point 5!

[201] Wood v Capita 41; *Pink Floyd Music Ltd and another v EMI Records Ltd.* [2010] EWCA Civ 1429 at [77]. *Credit Suisse Asset Management LLC v Titan Europe 2006-1 plc* [2016] EWCA Civ 1293, [28].

[202] *Arnold v Britton*, Lord Neuberger at [19, 20]

[203] In the UK, several standard form contracts are widely used in the construction and commercial property sectors, e.g. JCT Contracts (Joint Contracts Tribunal) NEC Contracts (New Engineering Contract) RIBA Contracts (Royal Institute of British Architects); in international shipping, the Baltic and International Maritime Council (BIMCO) provides standard contracts covering the full spectrum of shipping operations.

Let us consider a simple scenario where drafting starts with a template. Templates used in law firms often represent years of drafting expertise and, even if not all of them constitute the product of progressive refinements, they usually derive from contracts previously reviewed by senior lawyers.[204] While not necessarily "perfect," they are tailored to the jurisdiction and to the transaction at hand. As templates generally have a known lineage, both those who customize them and those who revise the customized document, can make certain assumptions as to their quality. This in turn facilitates review given that changes to a template are immediately visible and, as the document had already been "tried and tested," it need not be scrutinized afresh. In effect, the customization of templates provides more control over the drafting process and, assuming a known provenance, saves resources required to review. In contrast, generated contracts must be meticulously analyzed in their entirety because each of their provisions reproduces high-probability word sequences culled from millions of documents without regard to their substantive content, not to mention compatibility with other provisions. Generated contracts cannot be traced to a single document, source or author. No prompting technique can provide *any* assurances as to the quality of the generated text. In fact, it is generally impossible to determine *why* specific text was generated.[205] In many situations, however, the viability of a contract, be it generated or drafted, can only be evaluated if the reasoning underlying the formulation of individual provisions can be examined. Provisions that seem inadequate may, for example, constitute an unusual application of the law in light of the idiosyncrasies of a particular transaction provisions that seem plausible may turn out to be outside of the range of possible legal approaches. As indicated, determining the quality of contractual documents - or individual provisions - requires legal expertise and carries subjective undertones. Consequently, any evaluation of the generated output requires explainability. While text generation can be understood at a technical level as it is known *how* LLMs work, it is generally impossible to explain why a specific prompt resulted in a given output.[206] This lack of explainability derives from the non-deterministic nature of LLMs, the opacity of the training corpora and the aforementioned prompt sensitivity, amongst others. Admittedly, some LLMs disclose their "reasoning steps" and seem to provide *some* explainability.[207] Such explanations cannot, however, be trusted as LLMs are known to misrepresent their "reasoning" and to provide confident justifications for hallucinated text.[208] The inability to establish why a particular provision was generated is particularly troubling when users lack the legal expertise to verify the correctness of the generated text.

In sum, the customization of templates is not only more resource efficient than the generation of contracts, but also more transparent as it provides more control over the drafting process and the contents of the document. Apart from laborious prompt engineering, each contract generated by an LLM must be verified line by line as to its viability. To aggravate matters, the generated document often *appear* viable, creating a semblance of quality that may discourage thorough review and make mistakes, inconsistencies or even contradictions harder to detect.[209]

## Final Observations

Two contrasting narratives can be constructed: one enthusiastically advocates the use of LLMs for the generation of contracts and regards the generated documents as drafts that will undergo extensive review; the other bluntly declares that LLMs are useless. The first narrative underestimates the resources required to revise and rewrite the generated document. It also underestimates the resources required to litigate in order to "repair" the generated contract in situations where it was signed without review. It is, after all, statistically unlikely for an LLM to generate a viable contractual document that can be used without substantive amendments. The resources required to "repair" such document may exceed the resources required to draft a contract in a traditional manner. The second narrative overlooks the fact that LLMs can generate documents that not only look like contracts but, representing a brute-forced collage of popular contractual provisions, *could* contain the indispensable substance to be enforceable.

[204] Claire A. Hill, Why Contracts Are Written in "Legalese" (2001) 77 Chi.-Kent L. Rev. 59, 60, 62
[205] Sparks 60, 67; see also: Roman Yampolskiy, AI: Unexplainable, Unpredictable, Uncontrollable (2024, CRC Press).
[206] Søgaard A. On the opacity of Deep Neural Networks (2023) 53 Canadian Journal of Philosophy 224–239;
[207] Karthik Valmeekam et al., Planning in Strawberry Fields: Evaluating and Improving the Planning and Scheduling Capabilities of LRM o1 (October 3, 2024)
[208] Karthik Valmeekam et al., 7
[209] Emily M. Bender et al., On the Dangers of Stochastic Parrots: Can Language Models Be Too Big? ACM CONF. ON FAIRNESS, ACCOUNTABILITY, & TRANSPARENCY 610, 616–17 (2021) 617

It also fails to distinguish between questions of enforceability and questions of viability. Both narratives converge on the difficulties of (a) defining the minimum acceptable quality of what could be regarded as a viable contract, (b) quantifying the resources required to repeatedly prompt the LLM, review each generated output and amend the final version. Admittedly, with unlimited resources in terms of time, computation and legal as well as technical expertise on the side of their users, LLMs *could* generate viable contracts, not just contractual deepfakes. To recall: the viability of a contract is a function of its suitability to govern a specific transaction, evaluated in reference to its broader context, which includes the market conditions and the parties' bargaining positions.

We can endlessly debate whether - or to what extent – the combined principles of interpretation, implication and rectification can "repair" badly-generated contracts or whether courts should enforce commercially nonsensical bargains when the language of the contract is clear and its objective interpretation permits enforceability. Everything depends on the degree to which the generation had gone wrong and on the parties' willingness to litigate. It is one thing to blue-pencil a sentence or two, fix punctuation errors, fill in obvious gaps or rectify minor inconsistencies; it is yet another to reconcile many contradictory sentences, sever multiple provisions or fill in gaping omissions when the language of the contract precludes the determination of the parties' agreement.

For contracts governing idiosyncratic transactions requiring the reconciliation of multiple legal sources and/or novel legal appraoches, the use of LLMs is inadvisable. For contracts governing simple transactions, where an abundance of templates or suitable contractual documents already exist, the use of LLMs seems pointless. Why would anyone commit resources to generate a contract instead of customizing a template used on prior occasions? [210] It appears that the question is not whether LLMs *can* generate contracts but whether, in light of the resources required to review or "repair" them, whether they *should* be used for this purpose.

[210] J. Nyarko, Stickiness and Incomplete Contracts (2021) 88 U. Chi. L. Rev. 1